\documentclass[journal]{IEEEtran}

\usepackage{graphicx}
\usepackage{subfig}
\usepackage{amsmath,array}
\usepackage{amsfonts}
\usepackage{algorithm,algorithmic}
\usepackage{booktabs,multirow}
\usepackage{enumerate}
\usepackage{url}
\usepackage{hyperref}
\usepackage{color,soul}

\begin{document}

\title{Spatiotemporal Recurrent Convolutional Networks for Recognizing Spontaneous Micro-expressions}
\author{Zhaoqiang~Xia, Xiaopeng~Hong, Xingyu~Gao, Xiaoyi~Feng, Guoying~Zhao
\thanks{Z. Xia and X. Feng are with School of Electronics and Information, Northwestern Polytechnical University, 710129 Shaanxi, China. e-mail:xiazhaoqiang@gmail.com.}
\thanks{X. Gao is with Chinese Academy of Sciences, Beijing, China.}
\thanks{X. Hong and G. Zhao are with Center for Machine Vision and Signal Analysis, University of Oulu, 90014 Oulu, Finland.}
}

\markboth{arXiv}
{Z. Xia \MakeLowercase{\textit{et al.}}: Spatiotemporal Recurrent Convolutional Networks for Recognizing Spontaneous Micro-expressions}

\maketitle

\begin{abstract}
Recently, the recognition task of spontaneous facial micro-expressions has attracted much attention with its various real-world applications. Plenty of handcrafted or learned features have been employed for a variety of classifiers and achieved promising performances for recognizing micro-expressions. However, the micro-expression recognition is still challenging due to the subtle spatiotemporal changes of micro-expressions. To exploit the merits of deep learning, we propose a novel deep recurrent convolutional networks based micro-expression recognition approach, capturing the spatial-temporal deformations of micro-expression sequence. Specifically, the proposed deep model is constituted of several recurrent convolutional layers for extracting visual features and a classificatory layer for recognition. It is optimized by an end-to-end manner and obviates manual feature design. To handle sequential data, we exploit two types of extending the connectivity of convolutional networks across temporal domain, in which the spatiotemporal deformations are modeled in views of facial appearance and geometry separately. Besides, to overcome the shortcomings of limited and imbalanced training samples, temporal data augmentation strategies as well as a balanced loss are jointly used for our deep network. By performing the experiments on three spontaneous micro-expression datasets, we verify the effectiveness of our proposed micro-expression recognition approach compared to the state-of-the-art methods.
\end{abstract}


\IEEEpeerreviewmaketitle

\section{Introduction}
\IEEEPARstart{M}{icro-expressions} are very brief and involuntary facial expressions which are shown on the facial regions of humans. Compared with long-duration and obvious changes of normal facial expressions (namely macro-expressions), micro-expressions usually have short duration, i.e., less than $0.2$ second, and subtle intensity changes as facial muscle movements caused by micro-expressions only emerge in small and few regions \cite{zeng2009survey, shen2012effects}. The spontaneous micro-expressions can reveal the genuine emotions of humans and help understand humans' deceitful behaviors. Thus, it is potential to apply the micro-expressions in diverse fields \cite{Takalkar2018A}, such as lie detection, police case diagnosis, business negotiation, and psychoanalysis. Whereas, short duration and subtle changes of micro-expressions make it difficult for untrained people to detect and analyze micro-expressions. Even trained by professional micro-expression training tools \cite{ekman2007micro}, humans still manually detect and recognize micro-expressions from videos with low accuracy. Consequently, the automatic micro-expression recognition (MER) will be very valuable to promote the performance of analyzing large amounts of video sequences.

To tackle the MER task, several approaches have been presented to model subtle changes of micro-expressions in spatiotemporal domain \cite{Li2017Towards}. Most of these approaches are roughly divided into two main parts. The first part is to extract visual features from facial video clip towards MER. The second part is to choose a classifier for extracted features. Since MER is a typical pattern recognition task, some conventional classifiers, e.g., support vector machine (SVM) \cite{pfister2011recognising, Wang2015Micro, Huang2015Facial, Huang2016Spontaneous, Liong2018Less} and random forest \cite{pfister2011recognising, Davison2014Micro, Duan2016Recognizing, Li2017Towards}, have been used while the feature designing becomes more pivotal for solving MER problem in recent researches. Consequently, plenty of handcrafted features for macro-expressions or new-designed features have been explored in the past decade. For instance, the local binary patterns on three orthogonal planes (LBP-TOP) widely used to describe dynamic textures are firstly applied to recognize micro-expressions \cite{pfister2011recognising}. Although LBP-TOP has shown the capacity of discriminability and efficiency, it still suffers the sensitivity problem of global changes. So the second-order Gaussian jet on LBP-TOP \cite{Ruizhernandez2013Encoding}, LBP six intersection points (LBP-SIP) \cite{Wang2014LBP}, local spatiotemporal directional features (LSDF) \cite{Wang2015Rec}, spatiotemporal LBP (STLBP) \cite{Huang2015Facial}, spatiotemporal completed local quantization patterns (STCLQP) \cite{Huang2016Spontaneous}, directional mean optical-flow (MDMO) \cite{Liu2016A}, discriminative spatiotemporal LBP (DSLBP) \cite{Huang2017Discriminative} and bi-weighted oriented optical flow (Bi-WOOF) \cite{Liong2018Less} are proposed to improve the robustness of descriptors. These handcrafted features are designed to capture temporal differences of micro-expression sequences and achieve an accuracy  of more than $50\%$ \cite{Li2017Towards}.

However, it is still challenging to extract useful information from subtle changes and achieve high-quality descriptions as handcrafted features cannot well capture the subtle deformations of micro-expressions. Recently, deep convolutional neural networks (CNNs) have shown the great power in various fields and outperformed the handcrafted features as well as shallow classifiers \cite{Lecun2015Deep}. Deep learning approaches can obviate manual feature design and allow to automatically connect a specific task to the features themselves. Nevertheless, few deep models have been devoted to the MER problem due to limited video-based (sequence) training samples. Spontaneous micro-expression datasets usually contain insufficient samples, for instance, merely 256 micro-expression sequences for all categories in the largest CASME II dataset \cite{yan2014casme}. And they also have unbalanced classes, e.g., 26 sequences for category ``Happiness'' and 99 sequences for category ``Other'' in CASME II dataset. The limited and imbalanced samples will restrain deep CNNs as the deep network usually needs to learn numerous parameters. For alleviating this problem, the pre-trained CNN \cite{Takalkar2017Image} has been fine-tuned to recognize image based micro-expressions, in which each image (video frame) is assigned with a micro-expression category. The image based approach can obtain sufficient training samples by using video frames individually (rather than the entire video sequence) while the temporal changes are not considered. In order to leverage limited video-based samples for CNN based deep models, the temporal connectivity of CNNs which consider spatial and temporal changes jointly becomes vitally important for MER problem.

In this paper, we propose spatiotemporal recurrent convolutional networks (STRCN) to automatically recognize micro-expressions by conquering the ``limited and imbalanced training samples'' problem. To model the spatiotemporal motion deformations, we propose to employ CNNs with recurrent connections (i.e., recurrent convolutional networks) to learn the representation of subtle changes. The convolutional layers with recurrent connections are utilized to  learn visual features automatically and a classificatory layer is used to recognize micro-expressions. Towards the micro-expression video frames (clips), we exploit two types of extending the connectivity of STRCN across temporal domain for the network input. Moreover, to facilitate the learning procedure of deep model, we propose temporal augmentation strategies to greatly enrich the training samples for learning deep model and employ a balanced loss for counterweighing imbalanced classes.

Our main contributions are summarized as follows:
\begin{itemize}
  \item We propose an STRCN model to explore the powerful representation ability for MER problem. To the best of our knowledge, it is the first time that the spatiotemporal deformations are considered by deep models for MER problem.
  \item We propose two types of extending the connectivity of STRCN across temporal domain for video based micro-expressions, in which not only spatial information but also temporal changes are considered.
  \item We design some temporal augmentation strategies to greatly enrich the limited training samples for deep framework and employ a balanced loss for facilitating the imbalanced training.
\end{itemize}

The rest of this paper is organized as follows. Section \ref{sec:rw} reviews the related work briefly and our proposed deep framework for MER is presented in Section \ref{sec:pm}. Then we discuss the experimental results for algorithm evaluation in Section \ref{sec:ex}. Finally, Section \ref{sec:con} concludes this work.

\section{Related Work}
\label{sec:rw}
In this section, the researches on micro-expression analysis and deep learning for modeling spatiotemporal information are briefly summarized. The techniques for micro-expression analysis are described to indicate the shifted focus of research community while the deep learning studies are presented briefly to demonstrate the related techniques which can be used for MER problem.

\subsection{Micro-expression Analysis}
To date, some micro-expression datasets (Polikovsky’s dataset \cite{polikovsky2009facial}, USF-HD \cite{shreve2009towards}, SMIC \cite{li2013spontaneous}, CASME \cite{yan2013casme}, CASME II \cite{yan2014casme}, CASME\textsuperscript{2} \cite{Qu2017CAS} and SAMM \cite{Davison2018SAMM}) have been published in the literatures. Among them, the Polikovsky’s dataset \cite{polikovsky2009facial} and USF-HD dataset \cite{shreve2009towards} were constructed by collecting acted micro-expressions of subjects with high-speed cameras and are not available publicly. The remaining databases (SMIC, CASME, CASME II, CASME\textsuperscript{2} and SAMM) are spontaneous and obtained by collecting the induced micro-expressions of subjects watching specific videos. The CASME II is the extended version of CASME while CASME\textsuperscript{2} focuses on both micro-expressions and macro-expressions in long videos. Thus, in this paper, we study spontaneous MER problem using these three representative micro-expression datasets (i.e., SMIC \cite{li2013spontaneous}, CASME II \cite{yan2014casme}, and SAMM \cite{Davison2018SAMM}).

The task of spontaneous micro-expression analysis contains two subtasks: \emph{detection} and \emph{recognition}. The detection task is fundamental to subsequent recognition based on well-segmented video sequences containing micro-expressions while the recognition task aims to distinguish small differences between various kinds of micro-expressions. For the detection task, the geometric features \cite{shreve2011macro, Xia2016Spontaneous, Wang2017Main} and local textures \cite{moilanen2014spotting, Liong2016Spontaneous} have been proposed to capture micro-expression frames from videos. For the recognition task, several approaches \cite{Takalkar2018A} have been presented by using various features and classifiers, which are discussed in the following.

Some primary studies have been devoted to micro-expression recognition \cite{Li2017Towards}. In the earlier stage, scholars attempted to recognize acted micro-expressions. Polikovsky \textit{et al.} \cite{polikovsky2009facial} proposed a recognition method with descriptors of gradient orientation histogram on Polikovsky's dataset. In \cite{shreve2009towards}, authors presented the strain patterns for detecting macro-expressions and micro-expressions through the acted USF-HD dataset. However, the acted micro-expressions are greatly different from the spontaneous facial expressions \cite{afzal2009natural}. Therefore, recent works have further been done on the spontaneous facial micro-expressions, which can be roughly categorized as appearance based and geometric based methods.

\subsubsection{Appearance based Methods}
In recent studies, some appearance based features have been utilized to recognize micro-expressions. Pfister \textit{et al.} presented a recognition algorithm combing LBP-TOP features with SVMs, multiple kernel classifiers or random forests to recognize negative or positive micro-expressions \cite{pfister2011recognising}, in which the LBP-TOP features calculate the LBP features from three orthogonal planes and concatenate their histograms. Wang \textit{et al.} extended the LBP-TOP features into the tensor independent color space and then recognized micro-expressions in this subspace \cite{wang2014micro,Wang2015Micro}. However, the local textures described by LBP-TOP have the problems of robustness and sparse sampling \cite{Wang2014LBP, Huang2016Spontaneous}. Thus, some extended methods \cite{Wang2014LBP, Wang2015Rec, Huang2015Facial, Huang2016Spontaneous, Huang2017Discriminative} have further been proposed to improve the recognition performance. LBP-SIP provided a lightweight representation based on three intersecting lines crossing over the center point of LBP-TOP \cite{Wang2014LBP} and trained an SVM classifier. In \cite{Huang2015Facial, Huang2016Spontaneous}, the STCLBP and STCLQP features extended LBP-TOP features by containing more information (i.e., magnitude and orientation). Their hierarchical versions were reported in \cite{Zong2018Learning} and achieved better performance by considering using multiple blocks of LBP. In LSDF \cite{Wang2015Rec}, regions of interest (ROIs) were used to extract LBP-TOP features and further calculate local directional features, which encode the sign feature with magnitude information as weights. Furthermore, DSLBP \cite{Huang2017Discriminative} extracted 1D LBP and LBP-TOP combining with an integral projection and incorporates shape attributes into spatiotemporal texture features.

\subsubsection{Geometric based Methods}
Geometric-based methods extract deformation information from local landmarks or optical flow fields of facial regions without considering facial textures. In \cite{yao2014micro}, facial feature points have been tracked and used to recognize specific micro-expressions (i.e., happiness and disgust). Furthermore, the Delaunay triangulation based on extracted landmarks was used to reveal subtle muscle movements \cite{Lu2015Delaunay} and encoded temporally for dynamical micro-expressions. Besides, based on optical flow estimation, some approaches leverage the magnitude, orientation and other high-order statistics to model the dynamics of micro-expressions. The MDMO \cite{Liu2016A} features generated histograms from ROIs by counting the oriented optical flow vectors and then those histograms were used to recognize micro-expressions. Facial dynamics map (FDM) features calculated the optical flow direction based on the spatiotemporal cuboids and then were used to characterize a micro-expression sequence\cite{Xu2017Microexpression}. In Bi-WOOF \cite{Liong2018Less}, the orientation histogram in each block of facial regions was generated by considering the magnitude and optical strain values as the weights. The weighted histograms were further used for recognizing micro-expressions.

In this paper, we propose a deep framework implementing automatic feature learning for micro-expression recognition. Compared to these handcrafted features, our deep method can capture the subtle temporal changes automatically and obtain good recognition results by simultaneously learning a classifier. Besides, two types of temporal connectivities in our deep method model the spatiotemporal deformation in views of facial appearance and geometry separately.

\subsection{Spatiotemporal Deep Learning}
In recent years, CNNs have achieved great successes in many computer vision tasks and advanced the state-of-the-art accuracy of image/object recognition \cite{Urban2017Do, He2016Deep}. The architecture of CNNs is a type of artificial neural networks and has a purely feed-forward architecture characterized by local connections and weight sharing. Amounts of deep learning approaches based on CNNs have been developed to pursue more excellent performances by using millions of training samples. However, the conventional CNNs only capture the spatial variations without considering temporal information.

To model spatiotemporal variations, several deep approaches have exploited various architectures to be adaptive for video/sequence based data in many fields \cite{Herath2017Going}, such as video classification \cite{Karpathy2014Large, Du2015Learning, Srivastava2015Unsupervised} and action recognition \cite{Ji20133D, Wang2016Temporal, Zolfaghari2018ECO}, to cite only few. In \cite{Ji20133D}, the 3D convolutional filters were used to extract temporal information in every seven frames for action recognition. Similar to \cite{Ji20133D}, the size of filters on the first convolutional layer in the early fusion of \cite{Karpathy2014Large} was modified to some temporal extent, for example, $11\times11\times3\times10$, which can deal with 10 frames for one time. Similar idea has also been presented in \cite{Du2015Learning}, which can deal with full video frames at one time. In the late fusion of \cite{Karpathy2014Large}, two frames separately went through the same networks and merged in the first fully connected layer. These two ways can extract spatiotemporal information progressively in a dense sampling way. Different from the dense sampling strategy \cite{Ji20133D, Karpathy2014Large, Du2015Learning}, a sparse temporal sampling strategy in temporal segment network (TSN) \cite{Wang2016Temporal} was used to segment long-range video for enabling effective learning using the whole video. And more variants of TSN have been proposed to fuse mid/high-level features \cite{Zolfaghari2018ECO}. Besides, the long short-term memory based autoencoder model was introduced to learn compact representation for videos \cite{Srivastava2015Unsupervised}. More deep models based on aforementioned architectures have been proposed for modeling spatiotemporal information. These approaches have been proven to be effective for long-range videos and can well capture the significant changes. Whereas, it might not be effective and need to explore more specific models for the task containing subtle changes, such as micro-expression recognition.

\begin{figure*}[t]
  \centering
  \includegraphics[width=1.0\linewidth]{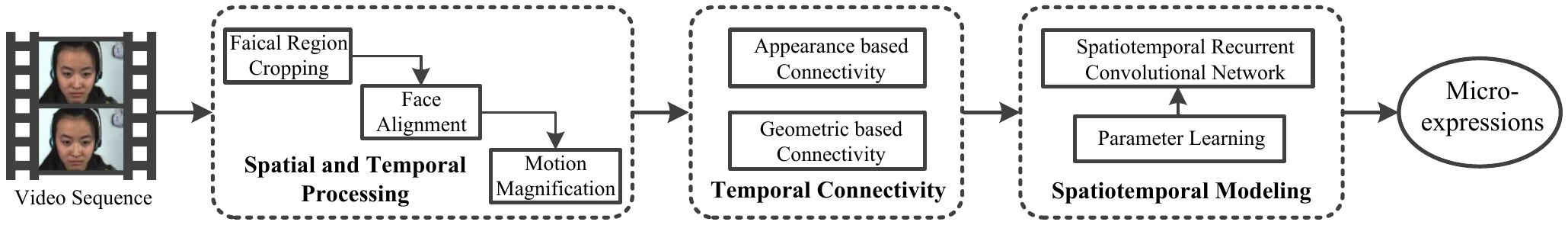}
  \caption{The framework of our proposed approach for micro-expression recognition.}
  \label{fig:frm}
\end{figure*}

More specifically, the feed-forward architectures can only capture the context in higher layers of CNNs where the units have larger receptive fields and fail to modulate the activities of units in lower layers. It means that only one-scale temporal variation can be captured by CNN based approaches. Inspired by neuroscience, the recurrent connections are adopted to obtain larger receptive fields and reduce the number of learnable parameters. The larger receptive fields are helpful to utilize the context information in lower layers of CNNs, and less parameters are more suitable for tasks without large amounts of samples \cite{Liang2015Recurrent}. In \cite{Pinheiro2014Recurrent} and \cite{Liang2015Recurrent}, two different types of convolutional layers with recurrent connections were proposed to label scenes. These recurrent convolutional networks were then applied to model significant changes for pain intensity regression based on static images \cite{Zhou2016Recurrent}. In this paper, we propose new spatiotemporal recurrent convolutional networks (STRCN) to model subtle changes for limited-range videos/sequences. The sampling strategies for long-range videos are not suitable for limited-range videos. To address this problem, we propose two types of extending the connectivity of STRCN across temporal domain for jointly modeling the spatial and temporal deformations.

\section{Proposed Method}
\label{sec:pm}
In this section, we will present our proposed deep model (i.e., spatiotemporal recurrent convolutional networks) for recognizing micro-expressions. To explain the details of our proposed STRCN method, total five parts will be successively introduced, including the framework, spatial and temporal processing, temporal connectivity, spatiotemporal modeling and parameter learning.

\subsection{The Framework}
The framework of our proposed method is shown in Fig. \ref{fig:frm}. Our proposed approach contains three key procedures, i.e., spatial and temporal processing, temporal connectivity and spatiotemporal modeling, for recognizing micro-expressions. For training deep models with limited and unbalanced samples, the parameter learning in training procedure is a key component of our proposed approach.

Firstly, we process video sequences of micro-expressions spatially and temporally. In spatial processing, the facial regions are cropped and aligned for each video (image sequence), resulting in the removal of non-facial regions. Then, in temporal processing, the motion deformations of facial regions are augmented to enhance subtle changes of micro-expressions. This will be introduced in Section \ref{sec:preproc} and an example is shown in Section \ref{sec:vip}. Secondly, two types of temporal connectivities are used to feed the sequential input into the subsequent deep model. The first type (denoted as \textbf{STRCN-A}) is an appearance based way while the second one (denoted as \textbf{STRCN-G}) is a geometric based way. This will be introduced in Section \ref{sec:tempcon}. At last, the spatiotemporal modeling with deep recurrent convolutional networks is used to recognize micro-expressions. The architectures and detailed setup will be introduced in Section \ref{sec:stmodel} and the parameter learning of deep models will be introduced in Section \ref{sec:paralearn}.

\subsection{Spatial and Temporal Processing}
\label{sec:preproc}
\subsubsection{Spatial Processing}
To avoid the pollution of regions without containing micro-expressions, facial regions need to be cropped and aligned from image sequences. In this context, the facial landmark points are detected firstly and then used to crop and align facial regions.

We employ an eye detector \cite{Xia2016An} and active shape model (ASM) algorithm \cite{Xia2016Spontaneous} to detect landmark points. The eye detector \cite{Xia2016An} can accurately locate two centering points of eye regions and determine the starting positions for face shapes described by ASM. And then the accurate locations of face shapes are iteratively fitted by the ASM algorithm \cite{Xia2016Spontaneous}. Following \cite{pfister2011recognising}, we use 68 landmark points to crop the facial regions. Given the centering points of eyes in frame $i$, i.e., $(x_i^l,y_i^l)$ for left eye and $(x_i^r,y_i^r)$ for right eye, the cropping coordinates of facial regions are calculated as
\begin{equation}
\begin{split}
topleft = ((x_i^l,y_i^l)) + \delta_1(0, y_i^l-y_i^r) - \delta_2(x_i^r-x_i^l, 0)  \\
height = \delta_3\sqrt{(x_i^l-x_i^r)^2+(y_i^l-y_i^r)^2} \\
width = \delta_4\sqrt{(x_i^l-x_i^r)^2+(y_i^l-y_i^r)^2}
\end{split}
\end{equation}

Based on the 68 landmark points, a local weighted mean (LWM) transformation \cite{pfister2011recognising} of any frame for sequence $i$ is used for aligning cropped facial regions. The transformed value of an arbitrary point is set to
\begin{equation}
f(x,y) = \frac{\sum_{i=1}^{N}W(\sqrt{(x-x_i)^2+(y-y_i)^2}/D_n)S_i(x,y)}{\sum_{i=1}^{N}W(\sqrt{(x-x_i)^2+(y-y_i)^2}/D_n)}
\end{equation}
where $W$ is the weight, $D_n$ is the distance of control point $(x_i,y_i)$ from its $(n-1)$th nearest control point, and $S_i(x,y)$ is the polynomial with $n$ parameters passing through a measure for $(x_i,y_i)$ and $n-1$ other measurements nearest to it. Using LWM transformation, all images in one sequence can be aligned frame by frame.

\subsubsection{Temporal Processing}
The temporal changes of aligned facial regions are very small and (almost) impossible to see with naked eyes of humans. Moreover, it is difficult to automatically learn representations of these subtle changes from noisy content by machine learning techniques. In this context, we utilize the motion magnification technique to amplify the hidden motion information of adjacent frames.

In this context, we choose the Eulerian Video Magnification (EVM) method \cite{Wu2012Eulerian} to amplify the temporal motion. The magnified temporal motion can be calculated by
\begin{equation}
\tilde{I}(x,y,t) = f(x,y) + \sum_k(1+\alpha_k)\delta_k(t)(\frac{\partial f(x,y)}{\partial x}, \frac{\partial f(x,y)}{\partial y})
\end{equation}
where $f(x,y) = I(x,y,0)$ and  $I(x,y,t)$ denotes the image intensity at position $(x,y)$ and time $t$. $\delta_k(t)$ is a displacement function and can be obtained by the temporal bandpass filter with respect to the $k$th frequency. $\alpha_k$ is a frequency-dependent motion magnification factor. $\tilde{I}(x,y,t)$ is the image intensity of $t$th frame after magnified.

In order to amplify the subtle changes of facial sequences, an infinite impulse response (IIR) filter is chosen as our temporal filter. In other words, we only use one temporal filter $\delta(t)$ ($k=1$) as the bandpass filter. After the temporal filtering, all images of each band are amplified with a fixed magnification factor $\alpha_k$. Finally, all bands of Laplacian pyramid are used to reconstruct the motion-magnified facial sequences.

\subsection{Temporal Connectivity}
\label{sec:tempcon}
\begin{figure}[t]
  \centering
  \includegraphics[width=0.8\linewidth]{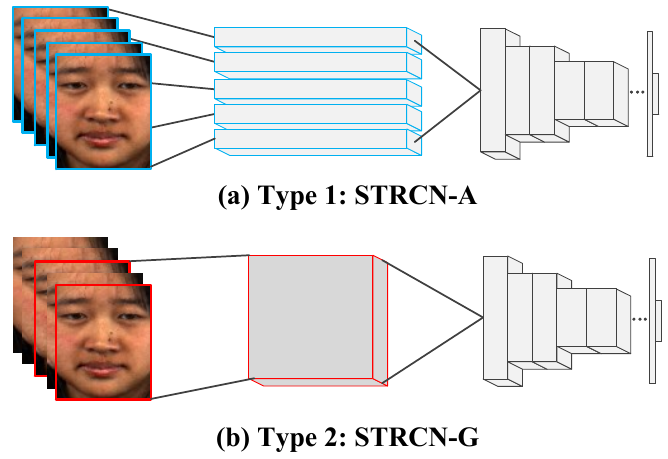}
  \caption{The schematic diagram of temporal connectivity. (a) Type 1: Appearance based connectivity (STRCN-A); (b) Type 2: Geometric based connectivity (STRCN-G).}
  \label{fig:tempconn}
\end{figure}
We investigate two types of temporal connectivities for fusing information across temporal domain. Type 1 concatenates all images in one sequence by vectorizing one channel of an image into a column of matrix. The spatiotemporal variations are reserved in a composite matrix and then appearance features can be learned by STRCN. So we call it as appearance based connectivity (abbreviated as \textbf{STRCN-A}). In type 2, the geometric motion is described by optical flow fields and then used to learn spatiotemporal features by STRCN. So we call it as geometric based connectivity (abbreviated as \textbf{STRCN-G}). The schematic diagram of these two types is shown in Fig. \ref{fig:tempconn}.

\subsubsection{Type 1: Appearance based Connectivity}
To obtain the appropriate input for convolutional networks, all pixels of entire image can be directly vectorized to a column of one matrix, however, this way may induce the lengthy size of matrix column. In STRCN-A, we propose a better way to reserve spatiotemporal information in one matrix. Usually, not all pixels in facial regions are helpful for learning representations. Observed from the magnified video sequences, micro-expressions are usually the fragments of normal expressions and only occur in particular areas of face, such as eyes, brows and mouth, which are called as \emph{micro-expression-aware areas} in this paper. Other facial areas, e.g., chin and hairs, do not reveal micro-expressions, which are \emph{micro-expression-unaware areas}. In addition, the short-duration micro-expressions do not involve wide-range facial changes. So, in the context, we choose to eliminate the effects of micro-expression-unaware facial areas and select out the micro-expression-aware ones. In other  words, only some useful pixels are chosen to be concatenated into a matrix.

In order to obtain micro-expression-aware areas, we propose a mask generation method and choose those areas with a mask. For generating a mask, a difference heat map based on entire datasets is calculated by accumulating temporal differences of video frames on entire datasets. The difference heat map can be further thresholded for generating the final mask. The difference heat map of magnified video frames can be calculated as follows:
\begin{equation}
\begin{split}
E(x,y) = \sum_i\sum_t d_i(x,y,t) \\
d_i(x,y,t) = |\tilde{I}_i(x,y,t) - \tilde{I}_i(x,y,0)|
\end{split}
\label{eq:heatmap}
\end{equation}
where $\tilde{I}_i(x,y,t)$ denotes the magnified image intensity at spatial position $x,y$ and time $t$ for $i$th facial sequence. $d_i(x,y,t)$ represents the difference of two frames with temporal interval $t$ and is accumulated to generate the heat map $E(x,y)$.

Through accumulating temporal differences, the micro-expression-aware facial areas become active in the difference heat map. We further design an efficient strategy to generate the final mask. Firstly, we sort all values of difference heat map in a descending order. Then the top $p\%$ percentiles of sorted values are chosen and others are abandoned. The binary mask is finally generated by setting the chosen values to $1$ and others to $0$, which has an example in Fig. \ref{fig:mask}. With the binary mask, the micro-expression-aware areas can be selected out and continue to be flattened to a tensor. Supposed that the magnified data $\tilde{I}(x,y,t)\in R^{d_1\times T}$ ($d_1 = W \times H$), the data after selection is denoted as $\hat{I}(x,y,t)\in R^{d_2\times T}$ ($d_2$ equals to the number of pixels in selected areas). Usually, the dimensionality $d_2$ after selection is greatly smaller than the original dimensionality $d_1$, i.e., $d_2 \ll d_1$. So, the entire video in STRCN-A can be denoted as a tensor $\mathcal{V}_1 \in \mathbb{R}^{d_1\times T \times C}$, where $C$ represents the image channels.

\begin{figure*}[t]
  \centering
  \includegraphics[width=0.95\linewidth]{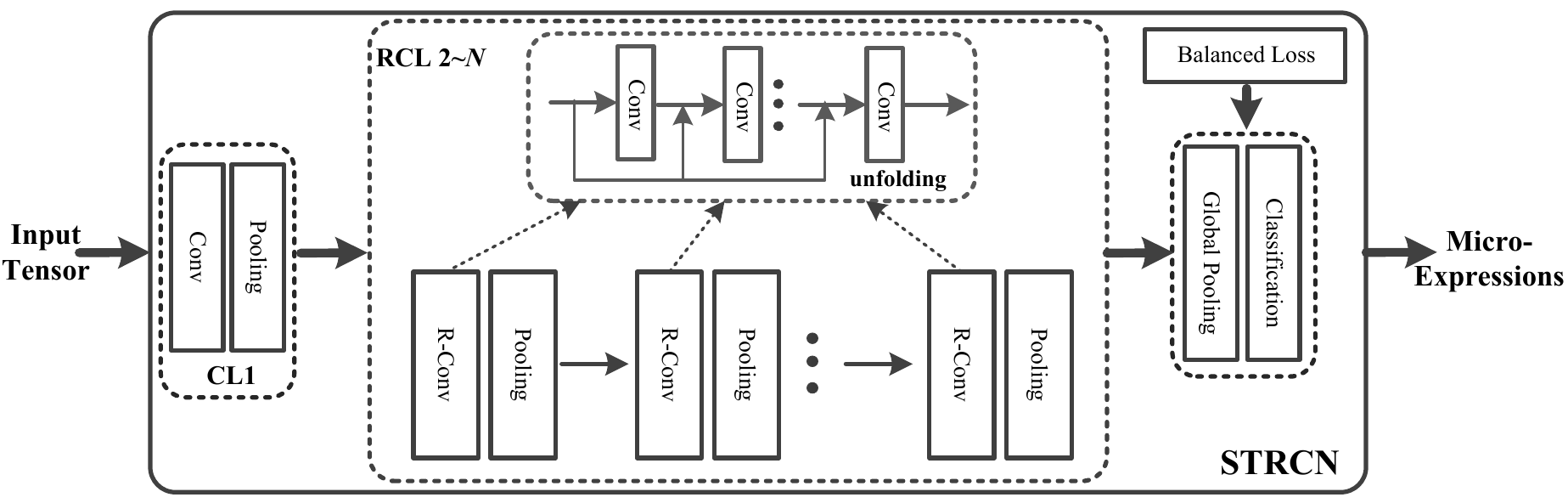}
  \caption{The architecture of deep RCNs for micro-expression recognition.}
  \label{fig:rcn}
\end{figure*}
\subsubsection{Type 2: Geometric based Connectivity}
In this type (STRCN-G), we investigate the motion information extracted from the entire video sequence by optical flow method. The dense sampling method used in \cite{Ji20133D,Karpathy2014Large, Du2015Learning} can model the motion variations of micro-expressions, however, it computes inefficiently and still need to further fuse the dense sampling. As the micro-expression clips are different from common video clips having multiple scenes and usually have one scene, we choose to use the sparse sampling to compute the optical flows. Inspired by \cite{Ngo2017Sparsity,Liong2018Less}, only the apex and onset frames are used to compute the optical flow images. Then the optical flow images are fed into STRCN for modeling geometric deformations.

The onset frame is defined as the first frame of video sequence containing micro-expressions while the apex frame is the frame containing the strongest-intensity expressions, which have the greatest changes from the first frame. To capture the greatest changes, we use the difference deviation to roughly locate the apex frame. The index $t_a$ of apex frame can be calculated as
\begin{equation}
\begin{split}
t_a = \underset{t}{\mathrm{arg\,max}} \, \sigma_i(t) \\
\sigma_i(t) = std\big(\tilde{I}_i(x,y,t) - \tilde{I}_i(x,y,0)\big)
\end{split}
\end{equation}
where $std(\cdot)$ computes the standard deviation of input matrix.

With the sparse sampling frames (i.e., onset and apex frames), we employ the accurate optical flow method \cite{Sun2010Secrets} for optical flow approximation. To compute more efficiently, the Lorentzian penalty function with improved model is chosen as the estimation method. The estimated displacement vector at position $(x,y)$ for sequence $i$ is denoted as $ (\mathbf{u}_{x,y},\mathbf{v}_{x,y}) $. So the entire video in STRCN-G can be denoted as a tensor $\mathcal{V}_2 \in \mathbb{R}^{W\times H \times 2}$.

\subsection{Spatiotemporal Modeling}
\label{sec:stmodel}
Compared to the handcrafted features, CNNs have more powerful ability to describe subtle changes of micro-expressions. In this paper, we add recurrent connections (i.e., RCNs \cite{Liang2015Recurrent}) within the feed-forward convolutional layers by using multiple-scale receptive fields. Based on two types of temporal connectivities, the spatiotemporal information of sequences can be further extracted by RCNs.

The architecture of our deep RCNs is shown in Fig. \ref{fig:rcn}. It contains one feed-forward convolutional layer (CL) and several recurrent convolutional layers (RCLs). The layer $1$ ($CL1$) is the only feed-forward convolutional layer without recurrent connections and used to compute efficiently. Following the standard convolutional layer 1, several RCLs ($RCL2\sim N$) are employed to extract visual features for recognition task. Between each convolutional layers (feed-forward and recurrent), max pooling operations are adopted to reduce dimensionality and save computation. Following the RCLs, a global average pooling layer is adopted to concatenate all feature maps into a vector. In the last layer, the classificatory layer with Softmax function is employed to calculate the recognition probabilities with concatenated feature vector.

In each RCL layer, several hidden layers are used to expand the size of receptive fields. One RCL layer can be unfolded into several convolutional layers. The layer latter in the subnetwork has larger receptive field in the same RCL layer. $R$ denotes the depth of one RCL layer, i.e., the number of hidden convolutional layers, and is valued from $1$ to $N$. For every convolutional layer, fixed-size feature maps are used to obtain the consistent connections. The input of an unit located at $(i,j)$ on the $k$th feature map in an RCL layer can be computed as
\begin{equation}
z_{ijk}(n) = {\mathbf{w}_k^f}^T \mathbf{u}_{ij}(n) + {\mathbf{w}_k^r}^T \mathbf{v}_{ij}(n-1)  + b_k
\end{equation}
where $\mathbf{u}_{ij}(n)$ and $\mathbf{v}_{ij}(n-1)$ represent the feed-forward and recurrent input, respectively. In the equation, $\mathbf{w}_k^f$ and $\mathbf{w}_k^r$ denote the feed-forward and recurrent weight vectors for $k$th feature map. $b_k$ is the bias of  $k$th feature map. The output of an unit located at $(i,j)$ on the $k$th feature map is given by
\begin{equation}
v_{ijk}(n) = f(z_{ijk}(n))
\end{equation}
where $n = 0,1,\cdots,N$ and the initial state $v_{ijk}(0)=0$. $f(\cdot)$ represents the normalized activation function.

Finally, the output of deep network uses the Softmax function to classify feature vectors to $C$ categories and it can be calculated as
\begin{equation}
\mathbf{p}_c = \frac{exp(\mathbf{W}^T\mathbf{v}) }{\sum_{c=1}^{C}exp(\mathbf{W}^T \mathbf{v})}
\end{equation}
where $\mathbf{p}_c$ is the predicted probability vector of all categories, $\mathbf{v}$ denotes the output feature vector of last pooling layer and $\mathbf{W}$ denotes the weight matrix between last global pooling layer and classification layer.

\subsection{Parameter Learning}
\label{sec:paralearn}
The parameter learning can be performed by minimizing the cross entropy loss function using the back propagation through time (BPTT) algorithm \cite{Liang2015Recurrent}. However, it is noted that two challenging issues for deep learning methods exist in current micro-expression datasets, which will make the parameter learning ineffective. The first one is having slight imbalanced classes while the second one is having limited samples. Take CASME II \cite{yan2014casme} for example. The category of ``Other'' have almost 4 times more samples than ``happiness''. Besides, no more than 250 original samples in CASME II \cite{yan2014casme} can be used to train deep RCNs. This will cause the problem of over-fitting and limit the recognition performance. To address these two problems, we employ two strategies, namely, multi-class balanced loss and multi-scale data augmentation to train the deep RCNs.

We extend the binary balanced loss \cite{Lin2018Focal} into MER problem with multiple categories. We define a quantity $\mathbf{p}_t $ as
\begin{equation}
\mathbf{p}_t = \mathbf{p}_c^{\mathbf{y}}(1-\mathbf{p}_c)^{1-\mathbf{y}}
\end{equation}
where $\mathbf{y}$ is the label vector and its arbitrary element $y_i \in\{0,1\}$. Then the multi-class balanced loss is computed as follows
\begin{equation}
\mathcal{L} = \sum_i\Big(-\beta_i log(\mathbf{p}_t^i) \Big)
\end{equation}
where $\beta_i$ is the weighting factor of class for sample $i$ and inversely proportional to the sample's class ratio in batch data. With the balanced loss function, the imbalance classes are balanced by $\beta_i$.

On the other side, we propose multi-scale data augmentation strategies to enrich training samples and further restrain the problem of small-size samples. Firstly, we use multiple-scale amplification factors $\alpha_k$ for training samples. All categories use multi-scale amplification factors $\alpha_k =[5, 14]$. With different factors, the size of samples are augmented by 10 times. Secondly, some frames are randomly selected out from one sequence with a percentage. Totally, five levels of percentages are adopted for random selection, i.e., $100\%$, $90\%$, $80\%$, $70\%$, and $60\%$. So, with five random selection, the data can be augmented by 5 times and these data contain different-sized sequences. Performing these two procedures jointly, the original data can be augmented by 50 times. These augmented data can make it sufficient for training deep architecture.

\section{Experiments}
\label{sec:ex}
In this section, we present the experimental details, including the implementation details, the datasets we used, the protocols, the approaches for comparison and experimental results.

\subsection{Implementation Details}
In the spatial processing procedure, the constants $\delta_1,\delta_2,\delta_3,\delta_4$ for cropping facial regions are set to $\{0.4,0.6,2.2,1.8\}$, following \cite{pfister2011recognising}. For the temporal processing, the cut-off frequencies of IIR filter are chosen as $[0.05, 0.4] Hz$. And the magnification factor in testing procedure is fixed to $\alpha = 8$. According to \cite{Wu2012Eulerian}, the bound for factor $\alpha$ in any frame is adopted as $\alpha_c = \frac{\lambda}{8\delta(t)}-1$. $\lambda$ denotes the spatial wavelength and is set to $\lambda = 16$ in this context. Therefore, the magnification factor can be finally used as $\alpha = min(8,\alpha_c)$.

In temporal connectivity procedure, the facial images are resized to fixed sizes for subsequent procedures. The fixed sizes are set to $64\times48$ for STRCN-A (type 1) and $300\times245$ for STRCN-B (type 2). For the mask generation in STRCN-A, considering the trade-off between the dimensionality reduction and information preserving, we choose the percentile value $p = 30\%$ as the threshold, which is further investigated in Section \ref{sec:pa}. The corresponding value of $70\%$ percentile of maximum is denoted as $E_p$ so that the positions with values $E(x)>E_p$ are selected as active elements. In the next step of STRCN-A, the temporal normalization operations are performed to obtain fixed-size input tensor for RCNs. Here, we choose $30$ frames to feed the deep model. For STRCN-B, the fixed-size ($300\times245$) frames are used to generate optical flow map and then fed to RCNs.

\begin{table}[t]
\centering
\caption{The detailed configuration of our deep STRCNs.}
\label{tab:rcn}
\scalebox{1.0}{
\centering
\begin{tabular}{|l|l|l|}
\hline
\multirow{2}{*}{Layers}	&\multicolumn{2}{|c|}{Configurations} \\  \cline{2-3}
		&\multicolumn{1}{ c|}{STRCN-A}	& \multicolumn{1}{ c|}{STRCN-G} \\  \hline \hline
Input	&Tensor:$d_1\times T\times3$	&Tensor: $W\times H\times2$ \\ \hline
Conv1	&\multicolumn{2}{|c|}{$k:5\times5$, $p:0$, $s:1\times1$}	\\ \hline 
Pool1	&MAX, $k:4\times1$, $s:4\times1$	&MAX, $k:4\times4$, $s:4\times4$	\\ \hline
\multirow{2}{*}{RCL2}	&\multicolumn{2}{|c|}{$1\; feed\text{-}forward$:  $k:1\times1$, $p:0$, $s:1\times1$} \\
		&\multicolumn{2}{|c|}{$3\; recurrents$:  $k:3\times3$, $p:1\times1$, $s:1\times1$}	\\ \hline
Pool2	&MAX, $k:4\times1$, $s:4\times1$	&MAX, $k:4\times4$, $s:4\times4$	\\ \hline
\multirow{2}{*}{RCL3}	&\multicolumn{2}{|c|}{$1\; feed\text{-}forward$:  $k:1\times1$, $p:0$, $s:1\times1$} 	\\
		&\multicolumn{2}{|c|}{$3 \; recurrents$:  $k:3\times3$, $p:1\times1$, $s:1\times1$}	\\ \hline
Pool3	&MAX, $k:4\times4$, $s:4\times4$	&MAX, $k:4\times4$, $s:4\times4$	\\ \hline
\multirow{2}{*}{RCL4}	&\multicolumn{2}{|c|}{$1\;feed\text{-}forward$:  $k:1\times1$, $p:0$, $s:1\times1$}	\\
		&\multicolumn{2}{|c|}{$3\;  recurrents$:  $k:3\times3$, $p:1\times1$, $s:1\times1$}	\\ \hline
Pool4	&MAX, $k:2\times2$, $s:2\times2$	&MAX, $k:2\times2$, $s:2\times2$	\\ \hline
\multirow{2}{*}{RCL5}	&\multicolumn{2}{|c|}{$1\;feed\text{-}forward$:  $k:1\times1$, $p:0$, $s:1\times1$}	\\
		&\multicolumn{2}{|c|}{$3 \; recurrents$:  $k:3\times3$, $p:1\times1$, $s:1\times1$}	\\ \hline
Pool5	&\multicolumn{2}{|c|}{Global AVG with alterable size} \\ \hline
Output	&\multicolumn{2}{|c|}{$C$ categories}	\\ \hline \hline
\multicolumn{3}{|l|}{All the convolutional layers contain $M$ feature maps.}	\\
\multicolumn{3}{|l|}{$k$ - filter or pooling size, $p$ - padding size, $s$ - stride size.}	\\ \hline
\end{tabular}}
\end{table}

The detailed configurations of RCNs are shown in Table \ref{tab:rcn}. Two types of temporal connectivities share the same architecture of RCNs but have different parameter setups. For each convolutional layer (feed-forward and recurrent), the batch normalization is used for scaling the activation and then a rectified linear unit is followed as the activation function. Since many parameters in our deep architecture may affect the performance of micro-expression recognition, we fix some parameters (e.g., filter sizes and stride size) with prior values in \cite{Liang2015Recurrent, Zhou2016Recurrent}. For another some important parameters (e.g., number of feature maps and recurrent layers), we explore their values by grid searching and investigate their impact in Section \ref{sec:pa}.

For learning parameters, the momentum is set to 0.9 and weight decay 0.0005 in stochastic gradient decent (SGD) procedure of BPTT. The stopping criterion for SGD is set to $10^{-3}$ for iterations. The learning rate is set to $10^{-3}$ in the beginning and will be multiplied with damping factor $0.8$ when all mini-batches are traversed and re-allocated randomly. To accelerate the parameter learning, we employ the library MatConvNet \cite{Vedaldi2015MatConvNet} to accomplish our proposed model. The mini-batch size for training model is set to $20$ as it is limited by the memory of GPUs (One Geforce TiTan X).

\subsection{Micro-expression Datasets and Setups}
Three representative micro-expression datasets are used to evaluate the performance of our proposed approach in our experiments: SMIC dataset \cite{li2013spontaneous}, CASME II dataset \cite{yan2014casme} and SAMM \cite{Davison2018SAMM}. All of them are specially designed to detect and recognize spontaneous micro-expressions, which are constructed by inducing subjects' micro-expressions. These three corpora have following characteristics:
\begin{itemize}
\item The SMIC dataset contains 164 spontaneous micro-expressions from 16 subjects. It is recorded by 100 fps high-speed cameras. These participants undergo high emotional arousal and suppress their facial expressions in an interrogation room setting with a punishment threat and highly emotional clips.
\item The CASME II dataset has 256 micro-expressions from 26 subjects. It has higher video quality and image size compared with SMIC. The recording rate of cameras in CASME II is 200 fps. Thus the video sequences of micro-expressions in CASME II have more frames than SMIC.
\item The SAMM dataset has 159 micro-expressions from 29 subjects at 200 fps. It uses similar procedures like CASME II but has a higher image resolution and employs an array of LEDs to avoid flickering. Because of the creators with professional rating skills, these expressions are obtained from stricter lab situations and labeled more accurately. And it has a wide ethnicity compared to other datasets.
\end{itemize}

To keep three datasets consistent with each other, we merge seven more categories in CASME II and SAMM into four classes. Following \cite{wang2014micro,Liu2016A,Zong2018Learning}, the happy micro-expressions in CASME II and SAMM are classified into ``Positive'' class as they indicate good emotions of subjects. In contrast, the disgust, sadness, fear, contempt and anger micro-expressions are classified into ``Negative'' class as they are usually considered as bad emotions. Surprise usually occurs when there is a difference between expectations and reality and can be neutral/moderate, pleasant, unpleasant, positive, or negative. Tense and repression are classified into the ``Other'' class as they indicate the ambiguous feelings of subjects and require further inference. In SMIC dataset, the first three classes (i.e., positive, negative and surprise) are used to annotate the micro-expressions.

For all experiments on three datasets, both \textit{leave-one-subject-out} (LOSO)  and \textit{leave-one-video-out} (LOVO) protocols are used to evaluate our proposed methods. To save the training time of deep models, we leave $5\%$ videos for testing, which will reduce the testing time. Based on two protocols, we use both accuracy and F1-score to evaluate the performance of our proposed STRCNs (STRCN-A and STRCN-G), avoiding the impact of imbalanced class problem for three datasets. Assume that $TP$, $FP$ and $FN$ are the true positive, false positive and false negative, respectively. The accuracy is calculated as $Acc = \frac{TP}{N}$ where $N$ is the number of testing samples. The F1-score is computed as $F=2\times \frac{P\times R}{P+R}$, where $P = \frac{\sum_{i=1}^{C} TP_i}{\sum_{i=1}^{C}(TP_i+FP_i)}$ and $R = \frac{\sum_{i=1}^{C} TP_i}{\sum_{i=1}^{C}(TP_i+FN_i)}$.

\subsection{Visualization of Intermediate Procedure}
\label{sec:vip}
\begin{figure}[t]
  \centering
  \includegraphics[width=0.85\linewidth]{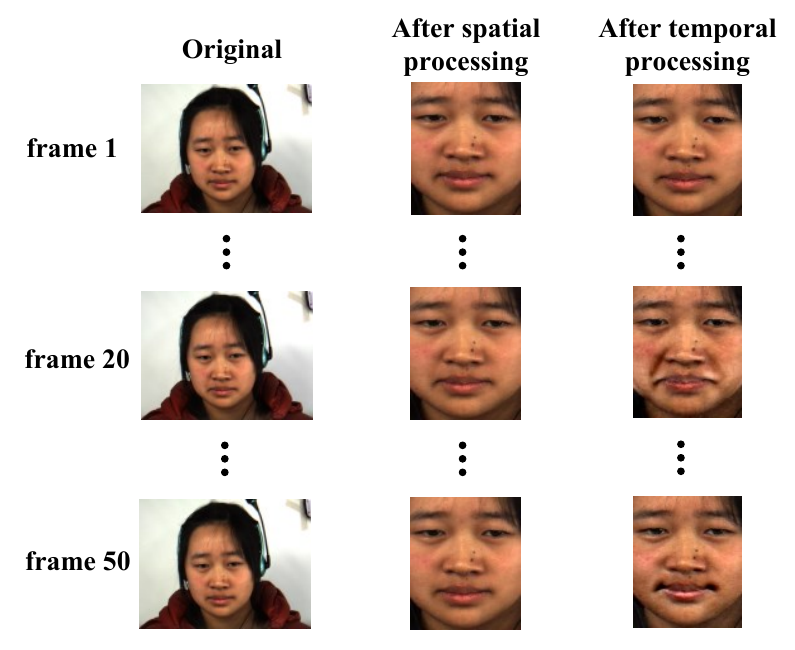}
  \caption{An example of using spatial and temporal processing (indexed ``sub02$\backslash$EP01\_11f'' in CASME II).}
  \label{fig:procexam}
\end{figure}
An example of spatial and temporal processing is shown in Fig. \ref{fig:procexam}, in which the magnification factor $\alpha$ is set to $8$. It is shown that the aligned facial regions can be obtained with spatial processing and the temporal changes are amplified by temporal processing. It is worth noticing that the temporal changes of micro-expressions cannot be observed easily by naked eyes without temporal processing (i.e., motion magnification), which also increases the difficulty of automatic learning. The temporal processing amplifies the temporal changes and helps to learn deep models.

\begin{figure}[t]
 \centering
  \includegraphics[width=0.75\linewidth]{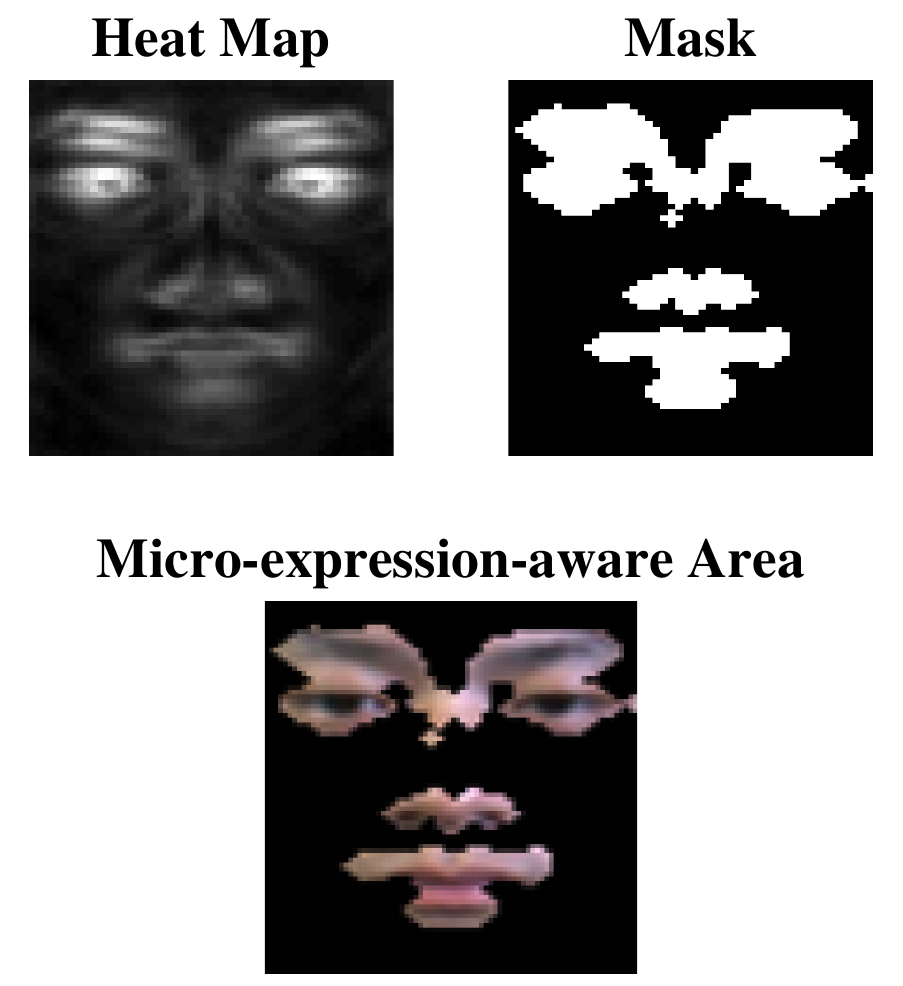}
  \caption{The difference heat map, generated binary mask and micro-expression-aware area ($p = 0.3$) in STRCN-A on entire datasets.}
  \label{fig:mask}
\end{figure}
The heat map and its corresponding mask in STRCN-A are visualized in Fig. \ref{fig:mask}. From the figure, it can be seen that some areas are not active in the difference heat map and might not reveal micro-expressions. Oppositely, the areas around eyes, nose and mouth are mostly active for micro-expressions and can be chosen by the binary mask. With the selected areas, the dimension of learning space for deep model can be drastically reduced and thus be helpful for the learning of deep model. The quantificational validation for selecting areas are investigated in \ref{sec:imper}.

\begin{table*}[t]
\centering
\caption{The recognition accuracy and F1-score of different methods under the \textbf{LOSO} protocol on three datasets.}
\label{tab:dm_loso}
\scalebox{1.0}{
\centering
\begin{tabular}{|c|c|c|c|c|c|c|}
\hline \hline
\multirow{2}{*}{Approaches}	&\multicolumn{2}{c|}{SMIC}	&\multicolumn{2}{c|}{CASME II}	&\multicolumn{2}{c|}{SAMM}  	\\ \cline{2-7}
                   		&Accuracy	&F1-score	&Accuracy	&F1-score	&Accuracy	 &F1-score \\
\hline \hline
LBP-TOP \cite{pfister2011recognising}	&0.457	&0.461 	&0.409	&0.369	&0.415 	 &0.406	\\ \hline
LBP-SIP \cite{Wang2014LBP}				&0.421	&0.422 	&0.457	&0.425	&0.417 	 &0.402	\\ \hline
LBP-TICS \cite{Wang2015Micro}			&0.439 	&0.384 	&0.405	&0.378	&0.395 	 &0.374	\\ \hline
STLBP-IP \cite{Huang2015Facial}			&0.543 	&0.547 	&0.551	&0.497	&0.568 	 &0.527	\\ \hline
Hierarchical STLBP-IP \cite{Zong2018Learning}			&0.601 	&0.613 	&0.638	 &0.611	&$N\backslash A$	&$N\backslash A$	\\ \hline
\hline
MDMO	\cite{Liu2016A}					&0.615&0.406	&0.510	&0.418	 &$N\backslash A$ &$N\backslash A$	\\ \hline
FDM	\cite{Xu2017Microexpression}	&0.714&0.540	&0.417	&0.297	&$N\backslash A$ &$N\backslash A$	\\ \hline
Bi-WOOF\cite{Liong2018Less}			&0.593&0.620	&0.589	&0.610	&0.598&0.591	 \\ \hline
\hline
Image-based CNN \cite{Takalkar2017Image}	&0.312	&0.305 	&0.444	&0.428	&0.436	 &0.429\\ \hline
\textbf{STRCN-A(Ours)}				&0.531	&0.514	&0.560	&0.542	&0.545	&0.492 \\ \hline
\textbf{STRCN-G(Ours)}				&\textbf{0.723}	&\textbf{0.695} 	 &\textbf{0.803}	&\textbf{0.747}	&\textbf{0.786} 	&\textbf{0.741}	\\
\hline \hline
\multicolumn{7}{r}{*$N\backslash A$ -  no results reported.}
\end{tabular}}
\end{table*}
\subsection{Comparison in LOSO Protocol}
Since the subject-independent evaluation protocol, i.e., LOSO, is becoming the main-stream for evaluating MER problem, we report the comparison results of our two deep models (STRCN-A and STRCN-G) in LOSO protocol with all state-of-the-art approaches, including the conventional appearance-based methods \cite{pfister2011recognising,Wang2014LBP,Wang2015Micro,Huang2015Facial,Zong2018Learning}, geometric-based methods \cite{Liu2016A,Xu2017Microexpression,Liong2018Less} and deep methods \cite{Takalkar2017Image} in Table \ref{tab:dm_loso}.

\subsubsection{Comparison Results to Appearance-based Methods}
The appearance-based methods are based on LBP and have LBP-TOP \cite{pfister2011recognising}, LBP-SIP \cite{Wang2014LBP}, LBP-TICS \cite{Wang2015Micro}, STLBP-IP \cite{Huang2015Facial} and Hierarchical STLBP-IP \cite{Zong2018Learning}. The LBP parameters of these methods are set as $R=3$ and $P=8$, which achieve the best performance in all configurations \cite{Wang2014LBP,Huang2015Facial,Zong2018Learning}. Except Hierarchical STLBP-IP, all LBP-based methods employ SVMs as classifiers, in which the polynomial kernel $k(x_i,x_j)= (\gamma_1 x_i^Tx_j+\gamma_2)^{\gamma_3}$ is used and the optimal values are set to $\gamma_1 = 0.22, \gamma_2 =0, \gamma_3 = 2$. All these methods are implemented according to their descriptions and retrained over three datasets. In contrast, the results with the optimal parameter set of kernelized group sparse learning are obtained from \cite{Zong2018Learning} for hierarchical STLBP-IP and only reported on two datasets, i.e., SMIC and CASME II.

Compared with the results of appearance-based methods in Table \ref{tab:dm_loso}, our proposed methods (STRCN-A and STRCN-G) achieve better results than them in most configurations and datasets. Especial for STRCN-G, its results are higher than all the other appearance based methods (including STRCN-A) under the LOSO protocol while STRCN-A achieve better performance only in LOVO protocol. In LOSO protocol, both our proposed appearance-based method (STRCN-A) and image-based CNN cannot achieve better performance than hierarchical STLBP-IP. That might be because the appearance-based deep models would be affected by the intra-class variations of each subject (person) as these deep models would learn certain appearance from each subject's samples.

\subsubsection{Comparison Results to Geometric-based Methods}
The geometric-based methods mainly utilize the optical flow fields to obtain the geometric information of facial movements and have MDMO \cite{Liu2016A}, FDM \cite{Xu2017Microexpression} and Bi-WOOF \cite{Liong2018Less}. The LOSO results originally from MDMO \cite{Liu2016A} and FDM \cite{Xu2017Microexpression} are used directly while the Bi-WOOF is implemented mainly according to \cite{Liong2018Less}. In Bi-WOOF, the block size for local weight and optical strain is set to $8 \times 8$. Then SVM classifiers are retrained on three datasets with the same parameters of LBP-based methods. Different from \cite{Liong2018Less}, the optical flow estimation method \cite{Sun2010Secrets} used in STRCN-G is also used to implement Bi-WOOF.

Compared with the results of geometric-based methods in Table \ref{tab:dm_loso}, our proposed geometric based method (STRCN-G) achieves better results than them in all configurations and datasets. Merely in SMIC dataset, FDM reports the accuracy near to our STRCN-G while FDM achieves very poor performance in CASME II dataset. For other approahces, STRCN-G achieves very promising performance improvement on all datasets. On the other side, in LOSO protocol, our proposed appearance based method (STRCN-A) only achieves better performance in some configurations and datasets. For instance, in SMIC dataset, all geometric based approaches achieve better accuracy and F1-score than the proposed STRCN-A under the LOSO protocol. Furthermore, the geometric-based methods achieve better performance than almost all appearance based methods, not just our proposed STRCN-A. Compared to appearance-based methods, geometric based methods may eliminate the intra-class information of each subject as only geometric information of subjects are reserved.

\subsubsection{Comparison Results to Deep Methods}
The image-based CNN \cite{Takalkar2017Image} recognizes the category of each frame in a micro-expression sequence. To compare with our proposed deep models fairly, we recognize one sequence as category $c_i$ when half frames in the sequence are recognized as the category $c_i$. Strictly following the image-based CNN, the VGGFace architecture is utilized to pre-train the CNN model\footnote{http://www.vlfeat.org/matconvnet/pretrained/} \cite{Vedaldi2015MatConvNet} and then fine-tuned in MER datasets. In \cite{Takalkar2017Image}, the temporal changes are not utilized for recognizing micro-expressions. In contrast, the temporal changes are modeled by two types of connectivities in our proposed methods.

Observed from Table \ref{tab:dm_loso}, our proposed methods outperform the image-based CNN obviously under the LOSO protocol. Even for the handcrafted features, the image based CNN cannot outperform them as those manually-designed features consider the temporal information of sequences. So it is indicated that the spatiotemporal modeling of deep models is pre-requisite for MER problem.

\begin{table*}[t]
\centering
\caption{The recognition accuracy and F1-score of different methods under the \textbf{LOVO} protocol on three datasets.}
\label{tab:dm_lovo}
\scalebox{1.0}{
\centering
\begin{tabular}{|c|c|c|c|c|c|c|}
\hline \hline
\multirow{2}{*}{Approaches}	&\multicolumn{2}{c|}{SMIC}	&\multicolumn{2}{c|}{CASME II}	&\multicolumn{2}{c|}{SAMM}  	\\ \cline{2-7}
                   		&Accuracy	&F1-score	&Accuracy	&F1-score	&Accuracy	 &F1-score \\
\hline \hline
LBP-TOP \cite{pfister2011recognising}	&0.618	&0.577 	&0.591	&0.537	&0.593 	 &0.542	\\ \hline
LBP-SIP \cite{Wang2014LBP}				&0.645	&0.598 	&0.657	&0.622	&0.645 	 &0.592	\\ \hline
LBP-TICS \cite{Wang2015Micro}				&0.637 	&0.586 	&0.643	&0.607	&0.647 	 &0.604	\\ \hline
STLBP-IP \cite{Huang2015Facial}				&0.701 	&0.669 	&0.723	&0.677	&0.712 	 &0.663	\\ \hline
Bi-WOOF \cite{Liong2018Less}				&0.745	&0.706	&0.725	&0.689	&0.718 	 &0.672	\\ \hline \hline
Image-based CNN \cite{Takalkar2017Image}	&0.531	&0.476 	&0.564	&0.507	&0.572	 &0.518\\ \hline
\textbf{STRCN-A(Ours)}				&\textbf{0.758}	&\textbf{0.714} 	 &\textbf{0.841}	&0.784	&\textbf{0.836} 	&\textbf{0.792} \\ \hline
\textbf{STRCN-G(Ours)}				&0.749	&0.710 	&0.833	&\textbf{0.807}	&0.827 	 &0.781	\\
\hline \hline
\end{tabular}}
\end{table*}
\subsection{Comparison in LOVO Protocol}
To complement the LOSO protocol, we also report the comparison results under the LOVO protocol. We compare our proposed methods with some conventional and deep methods on three datasets, including LBP-TOP \cite{pfister2011recognising}, LBP-SIP \cite{Wang2014LBP}, LBP-TICS \cite{Wang2015Micro}, STLBP-IP \cite{Wang2015Micro} Bi-WOOF \cite{Liong2018Less} and image-based CNN \cite{Parkhi2015Deep}. All these comparison methods are compared on three datasets and the performances are reported in Table \ref{tab:dm_lovo}. Among these state-of-the-art approaches, the variants of LBP method are based on appearance features while Bi-WOOF is a geometric based method. To compare deep models in LOVO protocol, we also compare our proposed method with image based CNN \cite{Takalkar2017Image}. Since we use $5\%$ of all samples, rather than only one sample, as the testing sample for each evaluation, all methods in  Table \ref{tab:dm_lovo} are reevaluated over three datasets with implementing their algorithms, rather than citing their reported results directly. The parameters of all methods are same to the configurations in LOSO protocol.

Observed from Table \ref{tab:dm_lovo}, our proposed methods (STRCN-A and STRCN-G) achieve the best performance in all configurations under the LOVO protocol. Compared to the handcrafted features, the deep features from automatic learning in our proposed method are competitive to other methods under the LOVO protocol. We can see that LBP-based features cannot outperform other methods as they are more suitable for the description of obvious changes of macro-expressions and the subtle changes can still not be well captured. Based on optical flow maps, the Bi-WOOF feature \cite{Liong2018Less} extracts subtle changes of micro-expressions but has limited describing ability. Our proposed methods can obtain the descriptions of subtle changes and outperform these state-of-the-art methods.

Moreover, the evaluation under the LOVO protocol is easier than LOVO protocol as all subjects can occur in training samples. Compared results in Table \ref{tab:dm_loso} and \ref{tab:dm_lovo}, our proposed methods, i.e., STRCN-A and STRCN-G, achieve different performances in both protocols. Especially, STRCN-A achieve better performance than STRCN-G in most configurations under the LOVO protocol. As STRCN-A is an appearance based method, which learns certain information of subjects, it can outperform STRCN-G in LOVO protocol while achieving poor performance in LOSO protocol.

\subsection{Parameter Analysis}
\label{sec:pa}
In this section, we analyze the parameters of the proposed methods and evaluate the impact of these parameters individually. Totally, the data augmentation, the balanced loss, the accuracy performance of the percentage threshold $p$, the size of feature maps and the number of recurrent layers are reported and discussed in this context.

\begin{table}[t]
\centering
\caption{The recognition accuracy of different methods without using data augmentation and balanced loss under the LOVO and LOSO protocol on three datasets.}
\label{tab:dabl}
\scalebox{1.0}{
\centering
\begin{tabular}{|c|c|c|c|}
\hline \hline
\multicolumn{4}{|c|}{LOVO}	\\ \hline
Approaches	&SMIC	&CASME II	&SAMM 	\\ \hline
STRCN-A without DA			&0.583	&0.593 	&0.617 \\ \hline
STRCN-G without DA			&0.605 	&0.605	&0.641 \\ \hline
STRCN-A without BL			&0.737	&0.829 	&0.831 \\ \hline
STRCN-G without BL			&0.741 	&0.818	&0.822 \\ \hline
\textbf{STRCN-A (Ours)}		&\textbf{0.758}	&\textbf{0.841} 	&\textbf{0.836} \\ \hline
\textbf{STRCN-G (Ours)}		&0.749	&0.833 	&0.827 \\
\hline \hline
\multicolumn{4}{|c|}{LOSO}	\\ \hline
Approaches	&SMIC	&CASME II	&SAMM 	\\ \hline
STRCN-A without DA			&0.481	&0.471 	&0.488 \\ \hline
STRCN-G without DA			&0.576 	&0.621	&0.642 \\ \hline
STRCN-A without BL			&0.541	&0.557 	&0.538 \\ \hline
STRCN-G without BL			&0.718 	&0.802	&0.775 \\ \hline
\textbf{STRCN-A (Ours)}		&0.531	&0.560 	&0.545 \\ \hline
\textbf{STRCN-G (Ours)}		&\textbf{0.723}	&\textbf{0.803} 	&\textbf{0.786} \\
\hline \hline
\end{tabular}}
\end{table}
\subsubsection{The Impact of Data Augmentation and Balanced Loss}
To observe the impact of temporal data augmentation and balanced loss, we replace the proposed methods with removing data augmentation (DA) procedure and balanced loss (BL). For removing the procedure of temporal data augmentation (DA), only the original samples are used by keeping model parameters consistent to proposed models. To fairly compare with final deep models (STRCN-A and STRCN-G), the same iterations are used in all deep models. For using imbalanced loss, we set all weights of samples to $\mathbf{\beta} = 1$ for removing the balancing weights. These methods are denoted as \textbf{STRCN-A without DA}, \textbf{STRCN-G without DA}, \textbf{STRCN-A without BL} and \textbf{STRCN-G without BL} in Table \ref{tab:dabl}.

The accuracy of proposed methods without DA and BL are reported in Table \ref{tab:dabl}. From the results, we can see that the temporal data augmentation can improve the performance of both two proposed deep models in both protocols. Obviously, the recognition ability of deep STRCN can be enhanced by leveraging more training samples. On the other side, the balanced loss can improve the performance slightly with considering the imbalanced classes. Overall, more sufficient training samples can promote the performances of deep models.

\begin{figure}[t]
  \centering
  \subfloat[LOVO]{\includegraphics[width=0.5\linewidth]{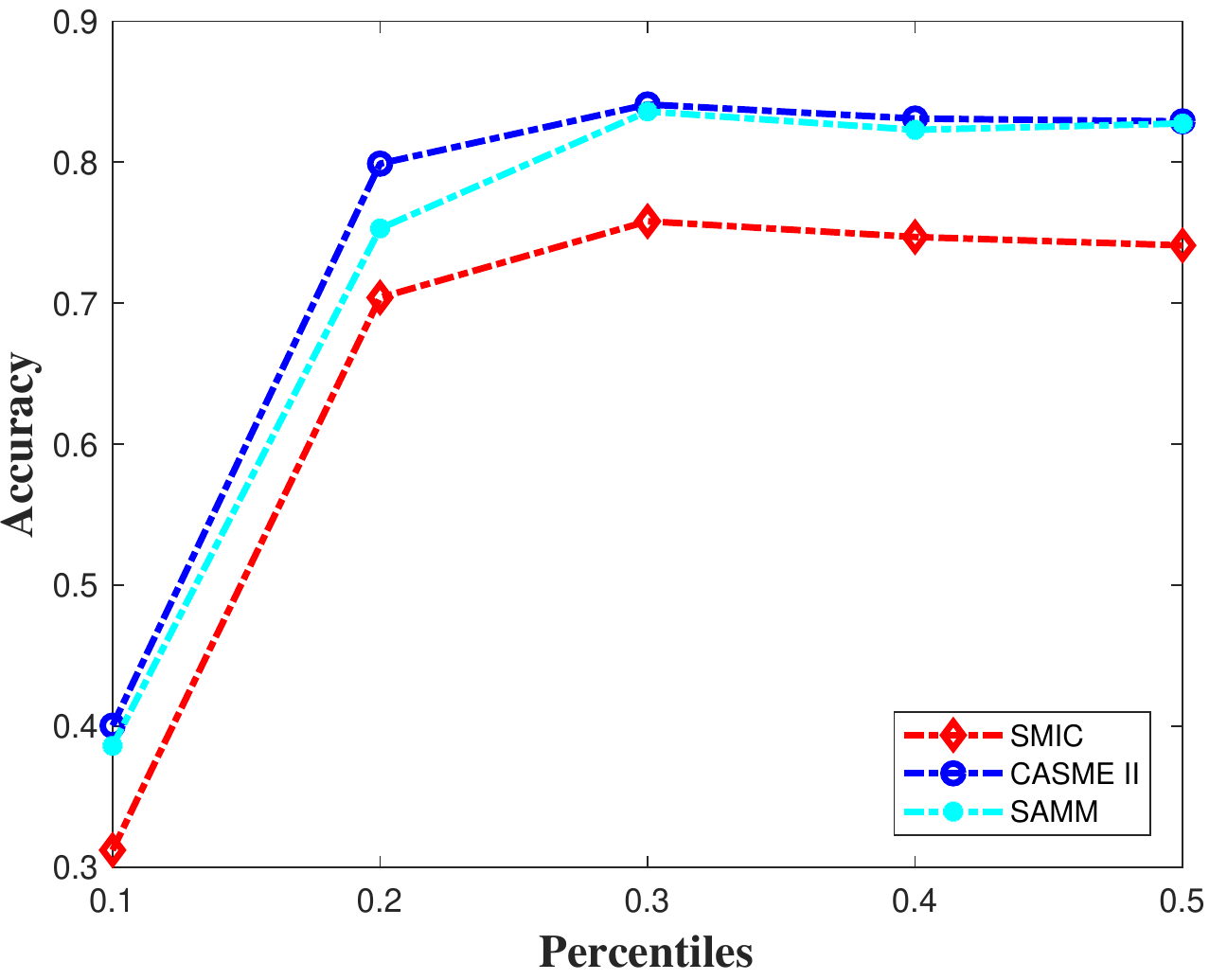}}
  \subfloat[LOSO]{\includegraphics[width=0.5\linewidth]{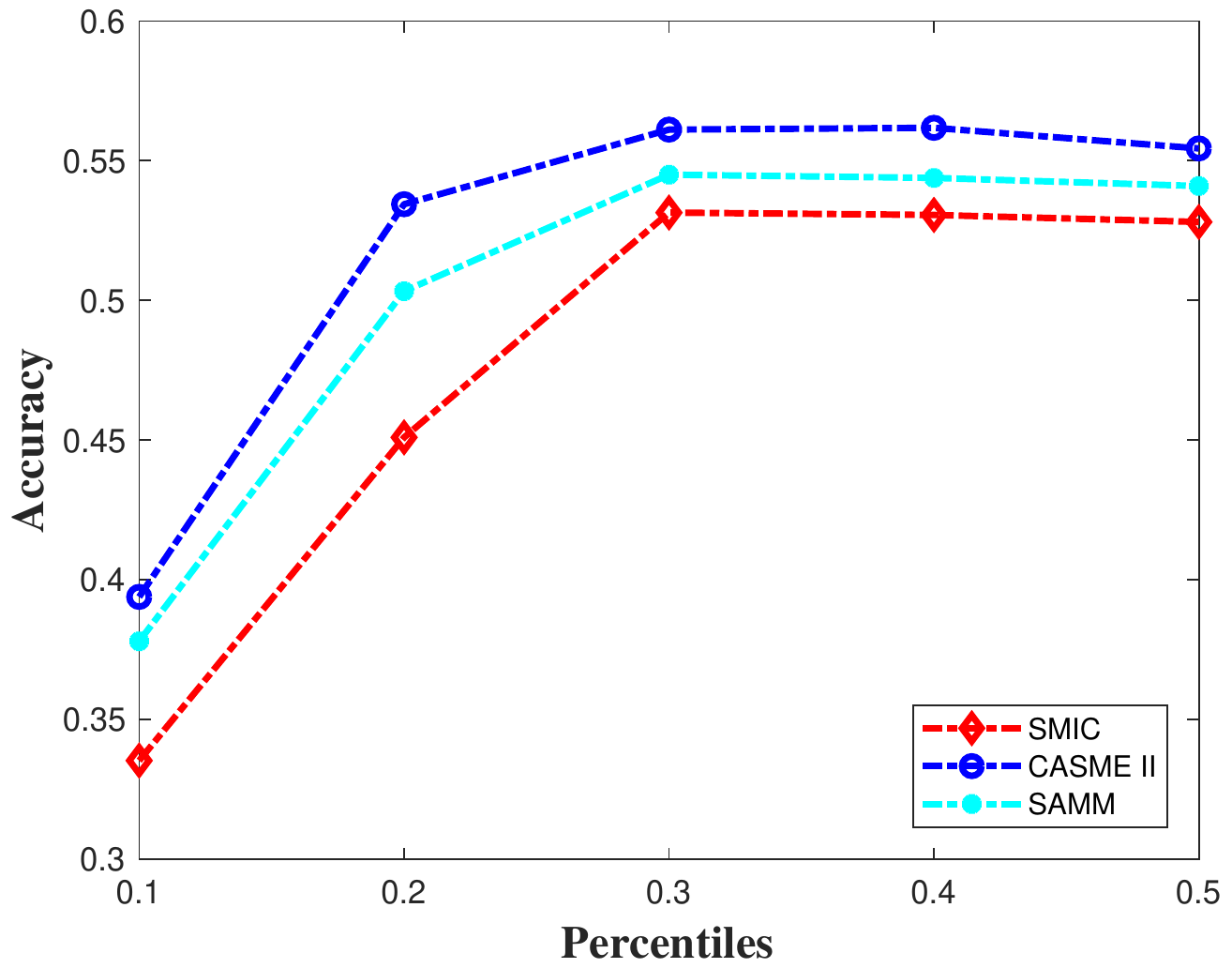}}
  \caption{The accuracy performance of various percentiles $p$ by STRCN-A on three datasets.}
  \label{fig:prctiles}
\end{figure}
\subsubsection{The Impact of Percentiles}
\label{sec:imper}
The values of $p$ in STRCN-A are evaluated by the accuracy of all datasets in both LOVO and LOSO protocols, which are shown in Fig. \ref{fig:prctiles}. According to the results, the performance will be degraded when too many pixels are abandoned by using lower values of $p$. In contrast, extra pixels may not be helpful for recognizing micro-expressions as some regions may contain noises. So the micro-expression-aware areas can be optimally determined by choosing appropriate values of $p$.

\begin{figure}[t]
  \centering
  \subfloat[LOVO, STRCN-A]{\includegraphics[width=0.5\linewidth]{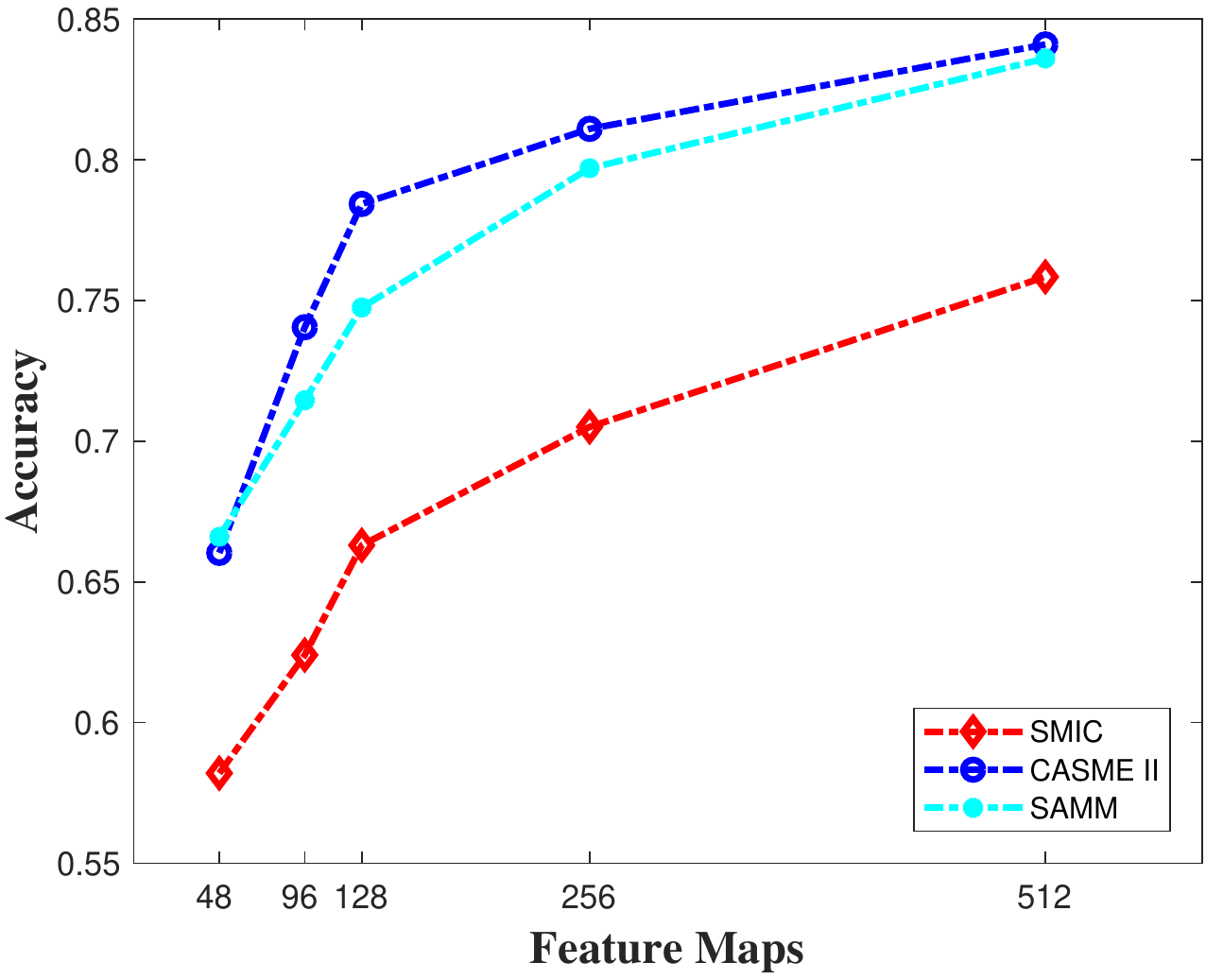}}
  \subfloat[LOSO, STRCN-A]{\includegraphics[width=0.5\linewidth]{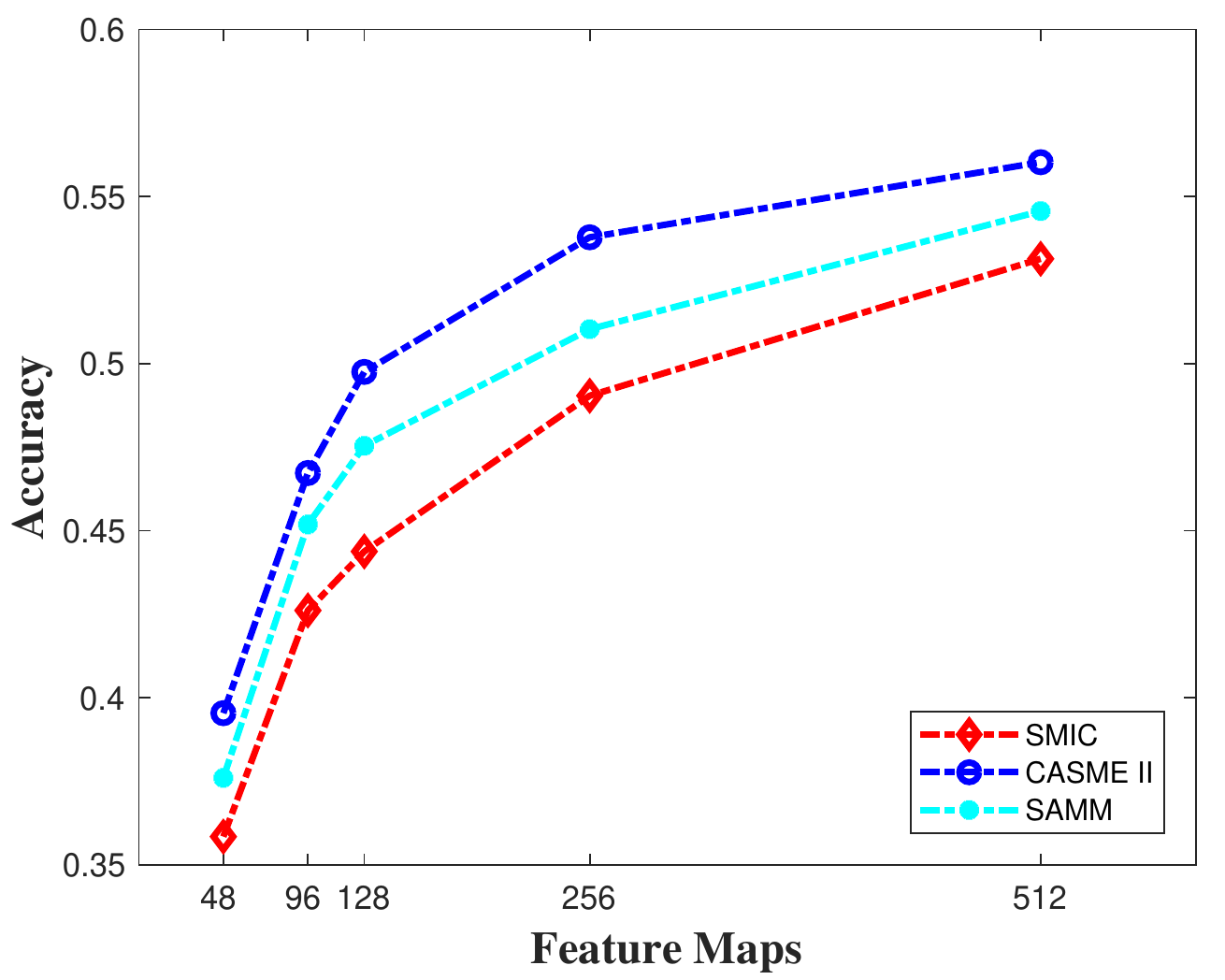}} \\
  \subfloat[LOVO, STRCN-G]{\includegraphics[width=0.5\linewidth]{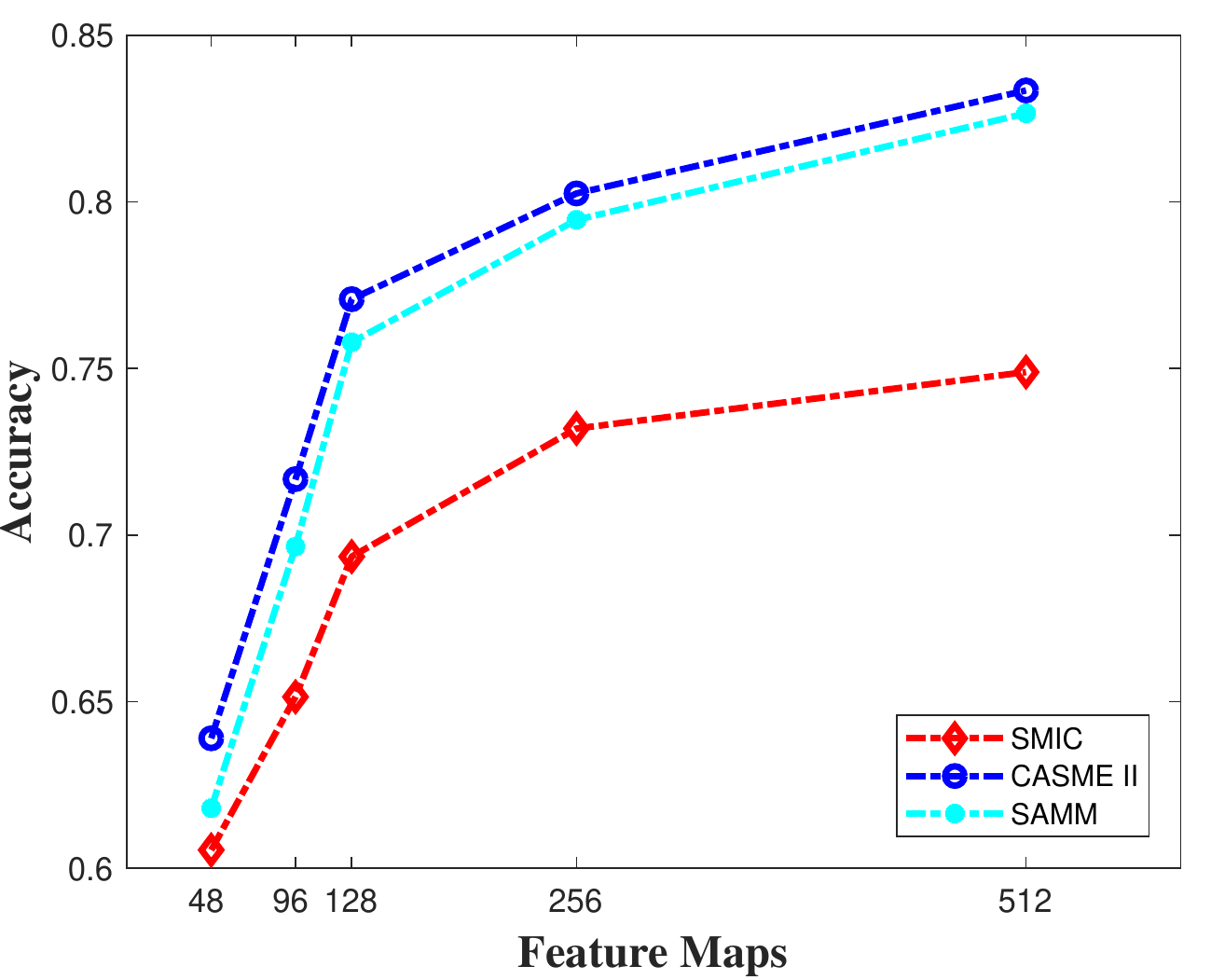}}
  \subfloat[LOSO, STRCN-G]{\includegraphics[width=0.5\linewidth]{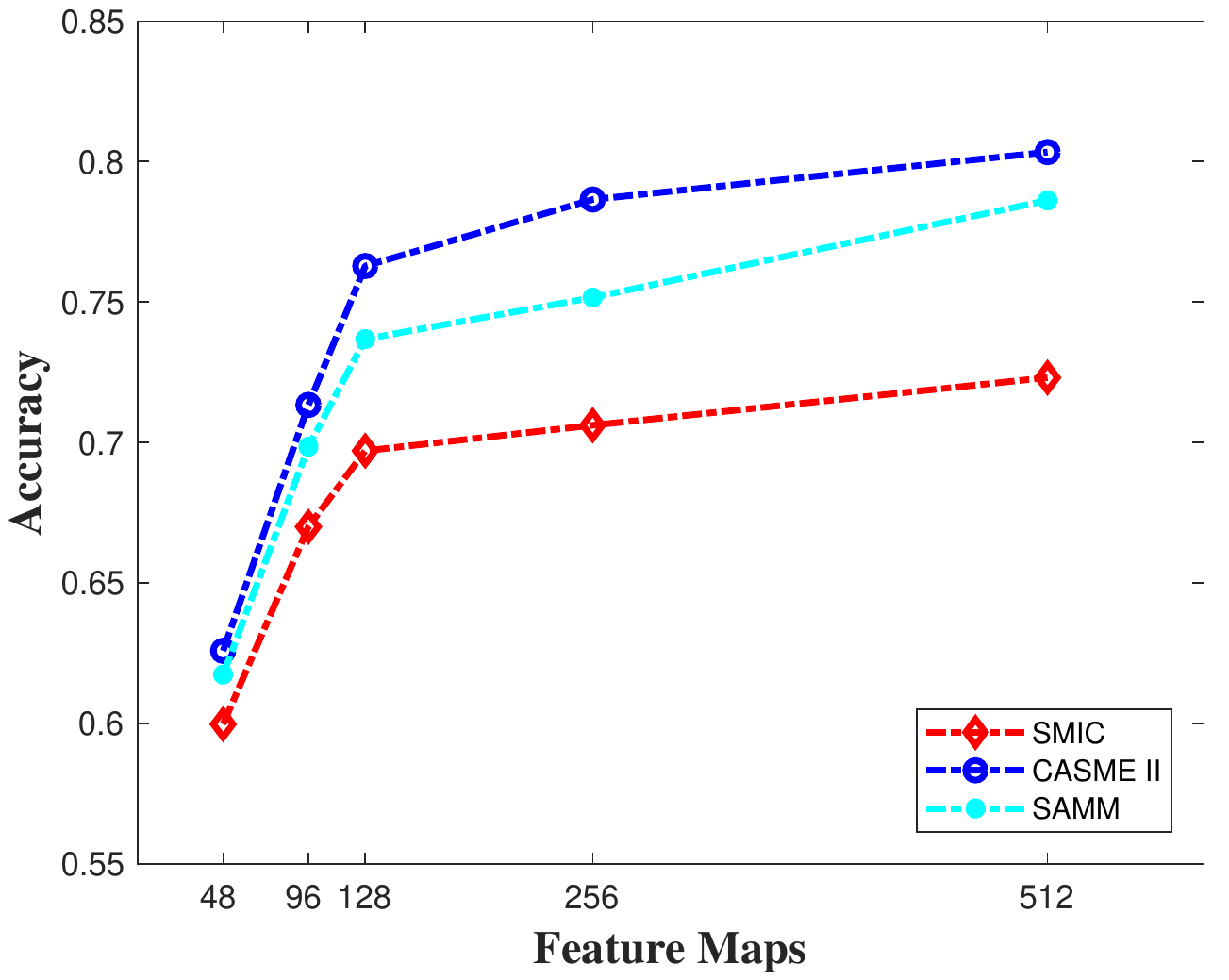}}
  \caption{The accuracy performance with different feature maps by Type-1 and Type-2 connections on three datasets.}
  \label{fig:fm}
\end{figure}
\subsubsection{The Impact of Feature Maps}
Figure \ref{fig:fm} shows the results of LOVO and LOSO evaluation for different feature maps on the SMIC, CASME II and SAMM datasets. It is noted that the performance can be steadily improved with more features maps on all datasets. More feature maps in each recurrent layer have more representation ability for extracting spatiotemporal information. However, the improvement will become less with the increase of feature maps. Besides, more feature maps occupy more memory and need more time to learn.

\begin{figure}[t]
  \centering
  \subfloat[LOVO, STRCN-A]{\includegraphics[width=0.5\linewidth]{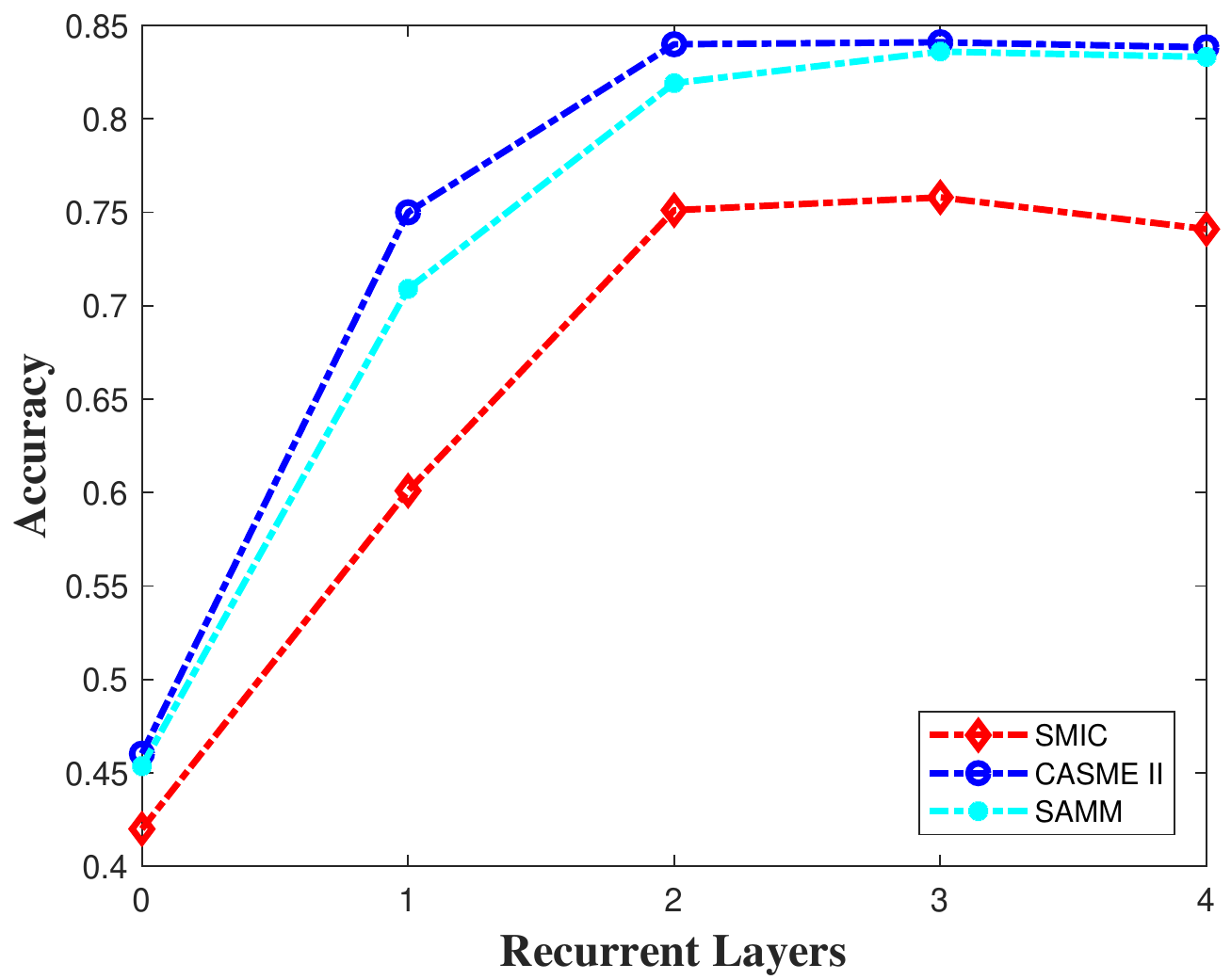}}
  \subfloat[LOSO, STRCN-A]{\includegraphics[width=0.5\linewidth]{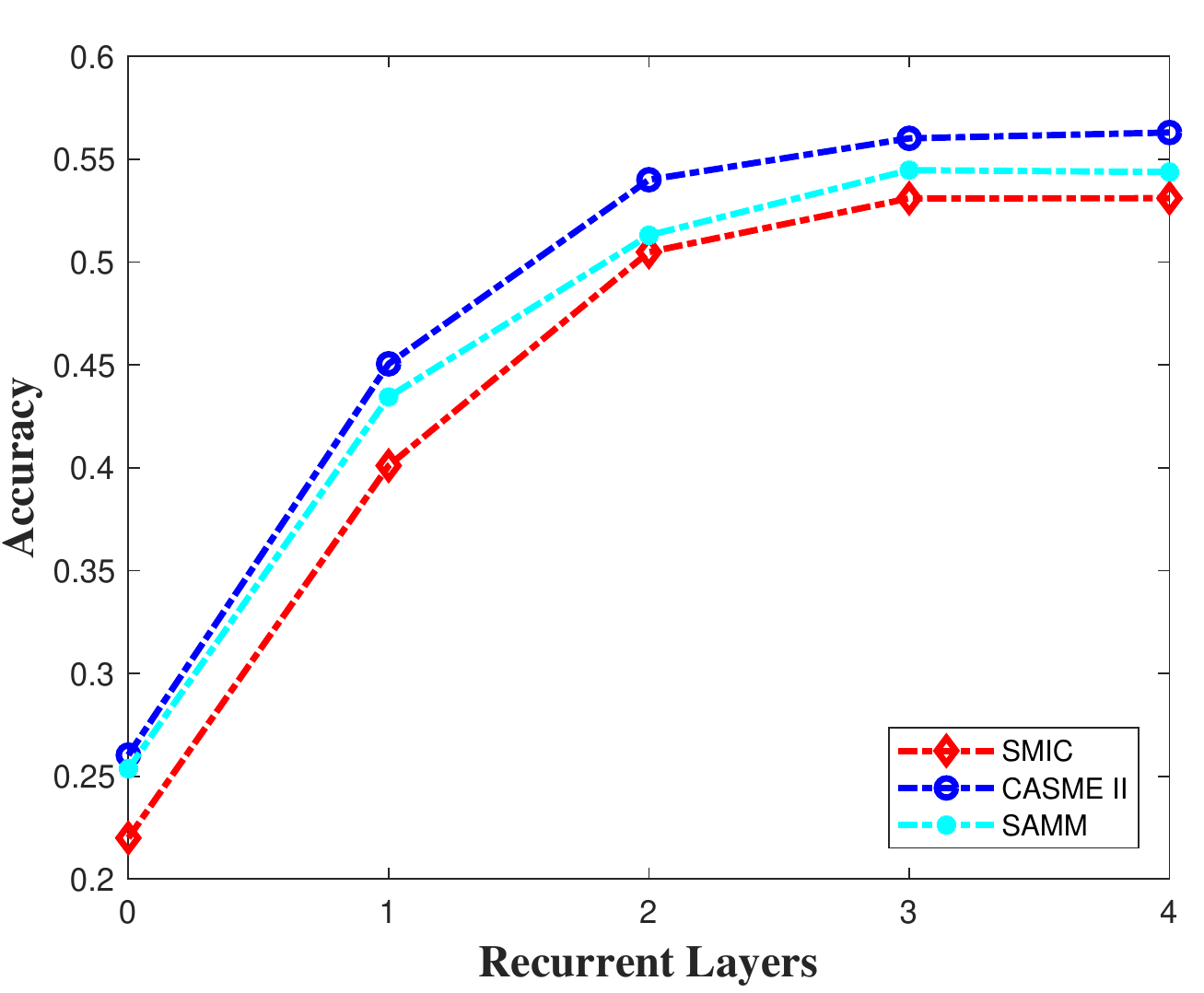}} \\
  \subfloat[LOVO, STRCN-G]{\includegraphics[width=0.5\linewidth]{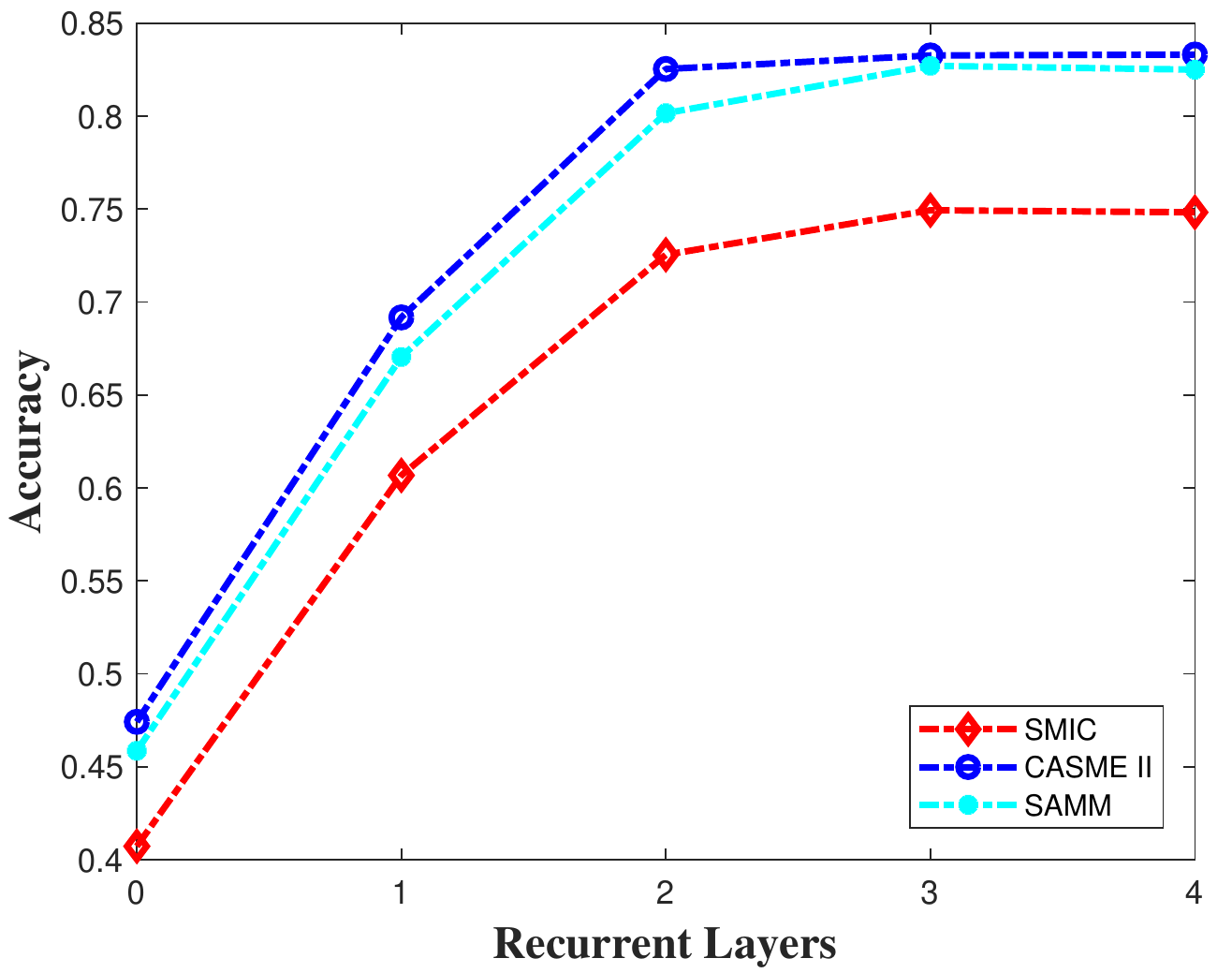}}
  \subfloat[LOSO, STRCN-G]{\includegraphics[width=0.5\linewidth]{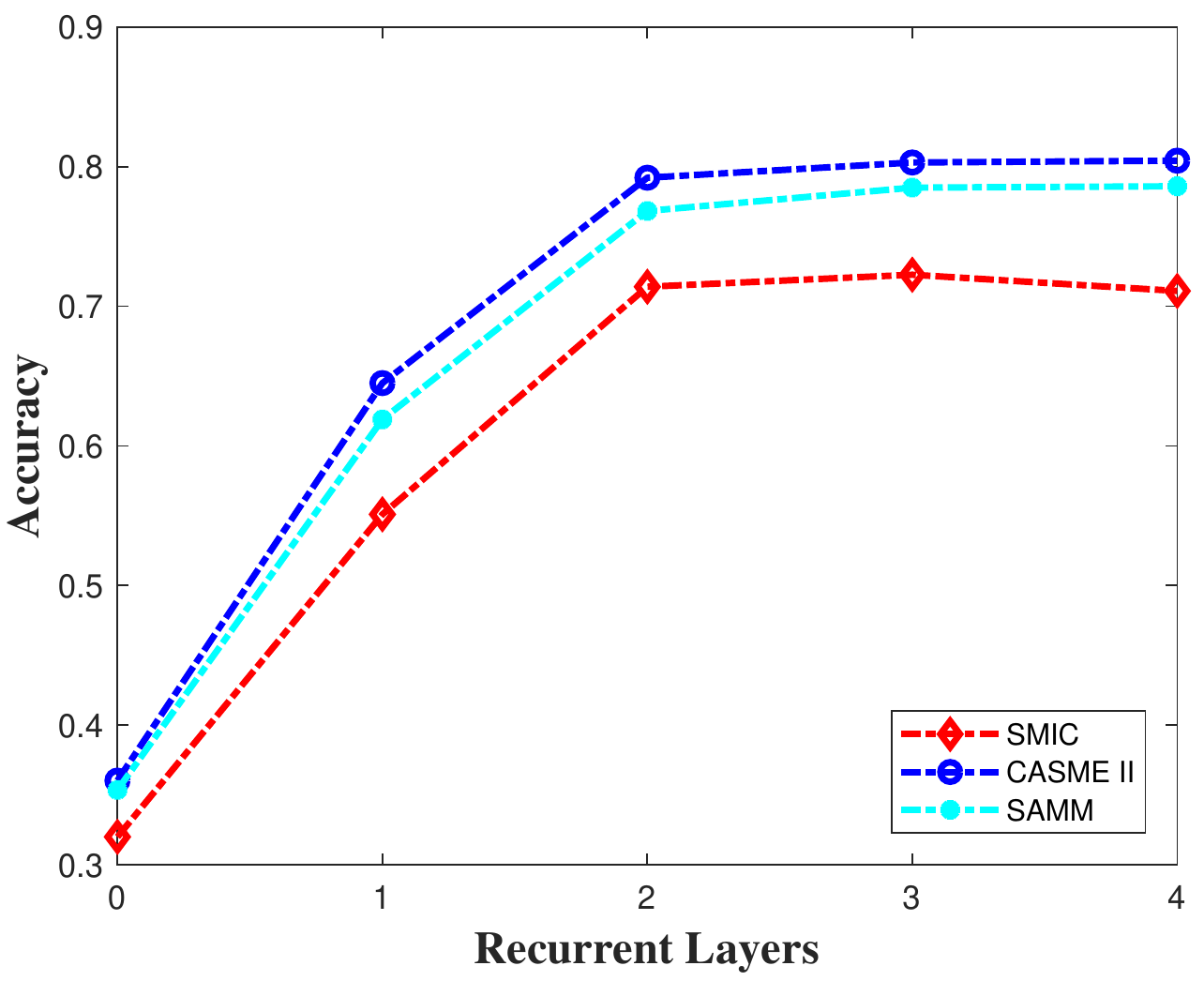}}
  \caption{The accuracy performance with different recurrent layers by STRCN-A and STRCN-G on three datasets.}
  \label{fig:rl}
\end{figure}

\subsubsection{The Impact of Recurrent Layers}
Different recurrent layers are evaluated on the SMIC, CASME II and SAMM datasets in Fig. \ref{fig:rl}. According to Table \ref{tab:rcn}, the RCL$2\sim5$ can be removed in sequence to obtain less recurrent layers (i.e., $1\sim4$ layers) while all recurrent layers can be replaced by convolutional layers to obtain zero recurrent layer (i.e., 0 layer). The experimental results show that more recurrent layers can improve the recognition performance. It is worth noting that the performance may decrease in some cases when there are too many recurrent layers, e.g., more than four layers in SMIC dataset. This might because the insufficient samples limit the performance as more recurrent layers need more training samples.

\section{Conclusion}
\label{sec:con}
In this paper, we proposed a novel micro-expression recognition approach based on spatiotemporal recurrent convolutional networks. The proposed STRCNs modeled the spatial-temporal deformations of micro-expression sequence by two types of connectivities (STRCN-A and STRCN-G), in which the connectivity of RCNs is extended across temporal domain for sequence based data. The STRCN-A was an appearance based method, which transformed one sequence into a matrix by concatenating frames to reserve the appearance of facial regions. In contrast, the STRCN-G was a geometric based method, which transformed one sequence into a matrix by computing the optical flow fields of onset and apex frames, to obtain the geometric information of facial movements. Furthermore, to overcome the shortcomings of limited and imbalanced training samples, temporal data augmentation strategies were designed for network input and a balanced loss was integrated to this recognition task. Through performing the experiments in LOVO and LOSO protocols on three spontaneous micro-expression datasets, i.e., SMIC, CASME II and SAMM, we verified the effectiveness of our proposed micro-expression recognition approach compared to the state-of-the-art methods. The STRCN-A achieved the best performance in the LOVO protocol while STRCN-G achieved the best performance in the LOSO protocol.

\section*{Acknowledgment}
This work is partly supported by the National Nature Science Foundation of China (Nos. 61702419, 61702491), the Natural Science Basic Research Plan in Shaanxi Province of China (No. 2018JQ6090), and the Tekes Fidipro Program of the Academy of Finland.

\bibliographystyle{IEEEtran}
\bibliography{IEEEabrv,refs}

\begin{thebibliography}{10}
\providecommand{\url}[1]{#1}
\csname url@samestyle\endcsname
\providecommand{\newblock}{\relax}
\providecommand{\bibinfo}[2]{#2}
\providecommand{\BIBentrySTDinterwordspacing}{\spaceskip=0pt\relax}
\providecommand{\BIBentryALTinterwordstretchfactor}{4}
\providecommand{\BIBentryALTinterwordspacing}{\spaceskip=\fontdimen2\font plus
\BIBentryALTinterwordstretchfactor\fontdimen3\font minus
  \fontdimen4\font\relax}
\providecommand{\BIBforeignlanguage}[2]{{%
\expandafter\ifx\csname l@#1\endcsname\relax
\typeout{** WARNING: IEEEtran.bst: No hyphenation pattern has been}%
\typeout{** loaded for the language `#1'. Using the pattern for}%
\typeout{** the default language instead.}%
\else
\language=\csname l@#1\endcsname
\fi
#2}}
\providecommand{\BIBdecl}{\relax}
\BIBdecl

\bibitem{zeng2009survey}
Z.~Zeng, M.~Pantic, G.~I. Roisman, and T.~S. Huang, ``A survey of affect
  recognition methods: Audio, visual, and spontaneous expressions,'' \emph{IEEE
  Transactions on Pattern Analysis and Machine Intelligence}, vol.~31, no.~1,
  pp. 39--58, 2009.

\bibitem{shen2012effects}
X.~Shen, Q.~Wu, and X.~Fu, ``Effects of the duration of expressions on the
  recognition of microexpressions,'' \emph{Journal of Zhejiang University
  SCIENCE B}, vol.~13, no.~3, pp. 221--230, 2012.

\bibitem{Takalkar2018A}
M.~Takalkar, M.~Xu, Q.~Wu, and Z.~Chaczko, ``A survey: facial micro-expression
  recognition,'' \emph{Multimedia Tools \& Applications}, vol.~77, no.~15, p.
  19301–19325, 2018.

\bibitem{ekman2007micro}
P.~Ekman, ``The micro-expression training tool, v. 2. (mett2),''
  www.mettonline.com, 2007.

\bibitem{Li2017Towards}
X.~Li, X.~Hong, A.~Moilanen, X.~Huang, T.~Pfister, G.~Zhao, and
  M.~Pietikäinen, ``Towards reading hidden emotions: A comparative study of
  spontaneous micro-expression spotting and recognition methods,'' \emph{IEEE
  Transactions on Affective Computing}, pp. 1--1, 2017.

\bibitem{pfister2011recognising}
T.~Pfister, X.~Li, G.~Zhao, and M.~Pietikainen, ``Recognising spontaneous
  facial micro-expressions,'' in \emph{International Conference on Computer
  Vision (ICCV)}.\hskip 1em plus 0.5em minus 0.4em\relax IEEE, 2011, pp.
  1449--1456.

\bibitem{Wang2015Micro}
S.~J. Wang, W.~J. Yan, X.~Li, and G.~Zhao, ``Micro-expression recognition using
  color spaces,'' \emph{IEEE Transactions on Image Processing}, vol.~24,
  no.~12, p. 6034, 2015.

\bibitem{Huang2015Facial}
X.~Huang, S.~J. Wang, G.~Zhao, and M.~Pietikäinen, ``Facial micro-expression
  recognition using spatiotemporal local binary pattern with integral
  projection,'' in \emph{ICCV Workshop on Computer Vision for Affective
  Computing}, 2015, pp. 1--9.

\bibitem{Huang2016Spontaneous}
X.~Huang, G.~Zhao, X.~Hong, W.~Zheng, and M.~Pietikäinen, ``Spontaneous facial
  micro-expression analysis using spatiotemporal completed local quantized
  patterns,'' \emph{Neurocomputing}, vol. 175, pp. 564--578, 2016.

\bibitem{Liong2018Less}
S.~T. Liong, J.~See, C.~W. Phan, and K.~S. Wong, ``Less is more:
  Micro-expression recognition from video using apex frame,'' \emph{Signal
  Processing: Image Communication}, vol.~62, pp. 82--92, 2018.

\bibitem{Davison2014Micro}
A.~K. Davison, M.~H. Yap, N.~Costen, K.~Tan, C.~Lansley, and D.~Leightley,
  ``Micro-facial movements: An investigation on spatio-temporal descriptors,''
  in \emph{European Conference on Computer Vision (ECCV)}, 2014, pp. 111--123.

\bibitem{Duan2016Recognizing}
X.~Duan, Q.~Dai, X.~Wang, Y.~Wang, and Z.~Hua, ``Recognizing spontaneous
  micro-expression from eye region,'' \emph{Neurocomputing}, vol. 217, pp.
  27--36, 2016.

\bibitem{Ruizhernandez2013Encoding}
J.~A. Ruizhernandez and M.~Pietikainen, ``Encoding local binary patterns using
  the re-parametrization of the second order gaussian jet,'' in
  \emph{International Conference and Workshops on Automatic Face and Gesture
  Recognition (FG Workshops)}, 2013, pp. 1--6.

\bibitem{Wang2014LBP}
Y.~Wang, J.~See, R.~C.~W. Phan, and Y.~H. Oh, ``{LBP} with six intersection
  points: Reducing redundant information in lbp-top for micro-expression
  recognition,'' in \emph{Proceedings of Asian Conference on Computer Vision
  (ACCV)}, 2014, pp. 21--23.

\bibitem{Wang2015Rec}
S.~J. Wang, W.~J. Yan, G.~Zhao, X.~Fu, and C.~G. Zhou, ``Micro-expression
  recognition using robust principal component analysis and local
  spatiotemporal directional features,'' in \emph{European Conference on
  Computer Vision Workshops (ECCV Workshops)}, 2015, pp. 325--338.

\bibitem{Liu2016A}
Y.~J. Liu, J.~K. Zhang, W.~J. Yan, S.~J. Wang, G.~Zhao, and X.~Fu, ``A main
  directional mean optical flow feature for spontaneous micro-expression
  recognition,'' \emph{IEEE Transactions on Affective Computing}, vol.~7,
  no.~4, pp. 299--310, 2016.

\bibitem{Huang2017Discriminative}
X.~Huang, S.~J. Wang, X.~Liu, G.~Zhao, X.~Feng, and M.~Pietikainen,
  ``Discriminative spatiotemporal local binary pattern with revisited integral
  projection for spontaneous facial micro-expression recognition,'' \emph{IEEE
  Transactions on Affective Computing}, vol.~PP, no.~99, pp. 1--1, 2017.

\bibitem{Lecun2015Deep}
Y.~Lecun, Y.~Bengio, and G.~Hinton, ``Deep learning,'' \emph{Nature}, vol. 521,
  pp. 436--44, 2015.

\bibitem{yan2014casme}
W.~J. Yan, X.~Li, S.~J. Wang, G.~Zhao, Y.~J. Liu, Y.-H. Chen, and X.~Fu,
  ``{CASME II}: An improved spontaneous micro-expression database and the
  baseline evaluation,'' \emph{PloS one}, vol.~9, no.~1, p. e86041, 2014.

\bibitem{Takalkar2017Image}
M.~A. Takalkar and M.~Xu, ``Image based facial micro-expression recognition
  using deep learning on small datasets,'' in \emph{International Conference on
  Digital Image Computing: Techniques and Applications (DICTA)}, 2017, pp.
  1--7.

\bibitem{polikovsky2009facial}
S.~Polikovsky, Y.~Kameda, and Y.~Ohta, ``Facial micro-expressions recognition
  using high speed camera and 3d-gradient descriptor,'' in \emph{International
  Conference on Crime Detection and Prevention}.\hskip 1em plus 0.5em minus
  0.4em\relax IET, 2009, pp. 1--6.

\bibitem{shreve2009towards}
M.~Shreve, S.~Godavarthy, V.~Manohar, D.~Goldgof, and S.~Sarkar, ``Towards
  macro-and micro-expression spotting in video using strain patterns,'' in
  \emph{Workshop on Applications of Computer Vision (WACV)}.\hskip 1em plus
  0.5em minus 0.4em\relax IEEE, 2009, pp. 1--6.

\bibitem{li2013spontaneous}
X.~Li, T.~Pfister, X.~Huang, G.~Zhao, and M.~Pietikainen, ``A spontaneous
  micro-expression database: Inducement, collection and baseline,'' in
  \emph{IEEE International Conference and Workshops on Automatic Face and
  Gesture Recognition (FG)}.\hskip 1em plus 0.5em minus 0.4em\relax IEEE, 2013,
  pp. 1--6.

\bibitem{yan2013casme}
W.~J. Yan, Q.~Wu, Y.-J. Liu, S.~J. Wang, and X.~Fu, ``{CASME} database: a
  dataset of spontaneous micro-expressions collected from neutralized faces,''
  in \emph{International Conference and Workshops on Automatic Face and Gesture
  Recognition (FG)}.\hskip 1em plus 0.5em minus 0.4em\relax IEEE, 2013, pp.
  1--7.

\bibitem{Qu2017CAS}
F.~Qu, S.~J. Wang, W.~J. Yan, H.~Li, S.~Wu, and X.~Fu,
  ``{CAS(ME)}\textsuperscript{2}: A database for spontaneous macro-expression
  and micro-expression spotting and recognition,'' \emph{IEEE Transactions on
  Affective Computing}, vol.~PP, no.~99, pp. 1--1, 2017.

\bibitem{Davison2018SAMM}
A.~K. Davison, C.~Lansley, N.~Costen, K.~Tan, and M.~H. Yap, ``{SAMM}: A
  spontaneous micro-facial movement dataset,'' \emph{IEEE Transactions on
  Affective Computing}, vol.~9, no.~1, pp. 116--129, 2018.

\bibitem{shreve2011macro}
M.~Shreve, S.~Godavarthy, D.~Goldgof, and S.~Sarkar, ``Macro-and
  micro-expression spotting in long videos using spatio-temporal strain,'' in
  \emph{International Conference on Automatic Face \& Gesture Recognition (FG)
  and Workshops}.\hskip 1em plus 0.5em minus 0.4em\relax IEEE, 2011, pp.
  51--56.

\bibitem{Xia2016Spontaneous}
Z.~Xia, X.~Feng, J.~Peng, X.~Peng, and G.~Zhao, ``Spontaneous micro-expression
  spotting via geometric deformation modeling,'' \emph{Computer Vision \& Image
  Understanding}, vol. 147, pp. 87--94, 2016.

\bibitem{Wang2017Main}
S.~J. Wang, S.~Wu, X.~Qian, J.~Li, and X.~Fu, ``A main directional maximal
  difference analysis for spotting facial movements from long-term videos,''
  \emph{Neurocomputing}, vol. 230, pp. 382--389, 2017.

\bibitem{moilanen2014spotting}
A.~Moilanen, G.~Zhao, and M.~Pietikainen, ``Spotting rapid facial movements
  from videos using appearance-based feature difference analysis,'' in
  \emph{International Conference on Pattern Recognition (ICPR)}.\hskip 1em plus
  0.5em minus 0.4em\relax IEEE, 2014, pp. 1722--1727.

\bibitem{Liong2016Spontaneous}
S.~T. Liong, J.~See, C.~W. Phan, Y.~H. Oh, A.~C.~L. Ngo, K.~S. Wong, and S.~W.
  Tan, ``Spontaneous subtle expression detection and recognition based on
  facial strain,'' \emph{Signal Processing Image Communication}, vol.~47,
  no.~C, pp. 170--182, 2016.

\bibitem{afzal2009natural}
S.~Afzal and P.~Robinson, ``Natural affect data—collection \& annotation in a
  learning context,'' in \emph{International Conference on Affective Computing
  and Intelligent Interaction and Workshops}.\hskip 1em plus 0.5em minus
  0.4em\relax IEEE, 2009, pp. 1--7.

\bibitem{wang2014micro}
S.-J. Wang, W.-J. Yan, X.~Li, G.~Zhao, and X.~Fu, ``Micro-expression
  recognition using dynamic textures on tensor independent color space,'' in
  \emph{International Conference on Pattern Recognition (ICPR)}.\hskip 1em plus
  0.5em minus 0.4em\relax IEEE, 2014, pp. 4678--4683.

\bibitem{Zong2018Learning}
Y.~Zong, X.~Huang, W.~Zheng, Z.~Cui, and G.~Zhao, ``Learning from hierarchical
  spatiotemporal descriptors for micro-expression recognition,'' \emph{IEEE
  Transactions on Multimedia}, vol.~PP, no.~99, pp. 1--1, 2018.

\bibitem{yao2014micro}
S.~Yao, N.~He, H.~Zhang, and O.~Yoshie, ``Micro-expression recognition by
  feature points tracking,'' in \emph{International Conference on
  Communications}.\hskip 1em plus 0.5em minus 0.4em\relax IEEE, 2014, pp. 1--4.

\bibitem{Lu2015Delaunay}
Z.~Lu, Z.~Luo, H.~Zheng, J.~Chen, and W.~Li, ``A delaunay-based temporal coding
  model for micro-expression recognition,'' in \emph{Asian Conference on
  Computer Vision Workshops (ACCV Workshops)}, 2015, pp. 698--711.

\bibitem{Xu2017Microexpression}
F.~Xu, J.~Zhang, and J.~Z. Wang, ``Microexpression identification and
  categorization using a facial dynamics map,'' \emph{IEEE Transactions on
  Affective Computing}, vol.~8, no.~2, pp. 254--267, 2017.

\bibitem{Urban2017Do}
G.~Urban, K.~J. Geras, S.~E. Kahou, O.~Aslan, S.~Wang, R.~Caruana, A.~Mohamed,
  M.~Philipose, and M.~Richardson, ``Do deep convolutional nets really need to
  be deep and convolutional?'' in \emph{International Conference on Learning
  Representations (ICLR)}, 2017, pp. 1--13.

\bibitem{He2016Deep}
K.~He, X.~Zhang, S.~Ren, and J.~Sun, ``Deep residual learning for image
  recognition,'' in \emph{International Conference on Computer Vision and
  Pattern Recognition}.\hskip 1em plus 0.5em minus 0.4em\relax IEEE, 2016, pp.
  770--778.

\bibitem{Herath2017Going}
S.~Herath, M.~Harandi, and F.~Porikli, ``Going deeper into action recognition:
  A survey,'' \emph{Image \& Vision Computing}, vol.~60, pp. 4--21, 2017.

\bibitem{Karpathy2014Large}
A.~Karpathy, G.~Toderici, S.~Shetty, T.~Leung, R.~Sukthankar, and F.~F. Li,
  ``Large-scale video classification with convolutional neural networks,'' in
  \emph{International Conference on Computer Vision and Pattern Recognition
  (CVPR)}.\hskip 1em plus 0.5em minus 0.4em\relax IEEE, 2014, pp. 1725--1732.

\bibitem{Du2015Learning}
T.~Du, L.~Bourdev, R.~Fergus, L.~Torresani, and M.~Paluri, ``Learning
  spatiotemporal features with 3d convolutional networks,'' in
  \emph{International Conference on Computer Vision (ICCV)}.\hskip 1em plus
  0.5em minus 0.4em\relax IEEE, 2015, pp. 4489--4497.

\bibitem{Srivastava2015Unsupervised}
N.~Srivastava, E.~Mansimov, and R.~Salakhutdinov, ``Unsupervised learning of
  video representations using lstms,'' in \emph{International Conference on
  Machine Learning (ICML)}, 2015, pp. 843--852.

\bibitem{Ji20133D}
S.~Ji, W.~Xu, M.~Yang, and K.~Yu, ``3d convolutional neural networks for human
  action recognition,'' \emph{IEEE Transactions on Pattern Analysis \& Machine
  Intelligence}, vol.~35, no.~1, pp. 221--231, 2013.

\bibitem{Wang2016Temporal}
L.~Wang, Y.~Xiong, Z.~Wang, Y.~Qiao, D.~Lin, X.~Tang, and L.~V. Gool,
  ``Temporal segment networks: Towards good practices for deep action
  recognition,'' in \emph{European Conference on Computer Vision (ECCV)}, 2016,
  pp. 20--36.

\bibitem{Zolfaghari2018ECO}
M.~Zolfaghari, K.~Singh, and T.~Brox, ``Eco: Efficient convolutional network
  for online video understanding,'' in \emph{European Conference on Computer
  Vision (ECCV)}, 2018, pp. 1--17.

\bibitem{Liang2015Recurrent}
M.~Liang and X.~Hu, ``Recurrent convolutional neural network for object
  recognition,'' in \emph{International Conference on Computer Vision and
  Pattern Recognition (CVPR)}.\hskip 1em plus 0.5em minus 0.4em\relax IEEE,
  2015, pp. 3367--3375.

\bibitem{Pinheiro2014Recurrent}
P.~H.~O. Pinheiro and R.~Collobert, ``Recurrent convolutional neural networks
  for scene parsing,'' in \emph{International Conference on Machine Learning
  (ICML)}, 2014, pp. 82--90.

\bibitem{Zhou2016Recurrent}
J.~Zhou, X.~Hong, F.~Su, and G.~Zhao, ``Recurrent convolutional neural network
  regression for continuous pain intensity estimation in video,'' in
  \emph{International Conference on Computer Vision and Pattern Recognition
  (CVPR) Workshops}.\hskip 1em plus 0.5em minus 0.4em\relax IEEE, 2016, pp.
  84--92.

\bibitem{Xia2016An}
Z.~Xia, W.~Zhang, F.~Tan, X.~Feng, and A.~Hadid, ``An accurate eye localization
  approach for smart embedded system,'' in \emph{International Conference on
  Image Processing Theory Tools and Applications (IPTA)}, 2016, pp. 1--5.

\bibitem{Wu2012Eulerian}
H.~Y. Wu, M.~Rubinstein, E.~Shih, J.~Guttag, F.~Durand, and W.~Freeman,
  ``Eulerian video magnification for revealing subtle changes in the world,''
  \emph{ACM Transactions on Graphics}, vol.~31, no.~4, pp. 13--15, 2012.

\bibitem{Ngo2017Sparsity}
A.~C.~L. Ngo, J.~See, and C.~W. Phan, ``Sparsity in dynamics of spontaneous
  subtle emotion: Analysis \& application,'' \emph{IEEE Transactions on
  Affective Computing}, vol.~8, no.~3, pp. 396--411, 2017.

\bibitem{Sun2010Secrets}
D.~Sun, S.~Roth, and M.~J. Black, ``Secrets of optical flow estimation and
  their principles,'' in \emph{International Conference on Computer Vision and
  Pattern Recognition (CVPR)}, 2010, pp. 2432--2439.

\bibitem{Lin2018Focal}
T.~Y. Lin, P.~Goyal, R.~Girshick, K.~He, and P.~Dollar, ``Focal loss for dense
  object detection,'' \emph{IEEE Transactions on Pattern Analysis \& Machine
  Intelligence}, vol. online, pp. 1--1, 2018.

\bibitem{Vedaldi2015MatConvNet}
A.~Vedaldi and K.~Lenc, ``Matconvnet - convolutional neural networks for
  matlab,'' in \emph{ACM Multimedia}.\hskip 1em plus 0.5em minus 0.4em\relax
  ACM, 2015, pp. 689--692.

\bibitem{Parkhi2015Deep}
O.~M. Parkhi, A.~Vedaldi, and A.~Zisserman, ``Deep face recognition,'' in
  \emph{British Machine Vision Conference (BMVC)}, 2015, pp. 1--12.

\end{thebibliography}

\end{document}